\documentclass[journal]{vgtc}                     
\onlineid{1693}
\vgtccategory{Area: Analytics \& Decisions}

\title{Visual Analysis of Multi-outcome Causal Graphs}

\author{%
  Mengjie Fan,
  Jinlu Yu,
  Daniel Weiskopf,
  Nan Cao,
  Huai-Yu Wang, and
  Liang Zhou*
}

\authorfooter{
  \item
  	M. Fan is with the Institute of Medical Technology, Peking University Health Science Center, and the National Institute of Health Data Science (NIHDS), Peking University.
  	E-mail: mengjiefan@bjmu.edu.cn.
   \item J. Yu is with Chalmers University of Technology and NIHDS, Peking University. E-mail: yjl.intp@gmail.com
  \item
        L. Zhou is with NIHDS, Peking University. L. Zhou is the corresponding author.
  	E-mail: zhoulng@pku.edu.cn.
   \item D. Weiskopf is with the Visualization Research Center (VISUS), University of Stuttgart.
        E-mail: weiskopf@visus.uni-stuttgart.de
    \item N. Cao is with the iDV$^{x}$ Lab, Tongji University.
        E-mail: nan.cao@gmail.com
    \item H. Wang is with the National Institute of Traditional Chinese Medicine Constitution and Preventive Treatment of Diseases, Beijing University of Chinese Medicine.
        E-mail: wanghuaiyuelva@126.com
}

\abstract{%
  We introduce a visual analysis method for multiple causal graphs with different outcome variables, namely, multi-outcome causal graphs.
  Multi-outcome causal graphs are important in healthcare for understanding multimorbidity and comorbidity.
  To support the visual analysis, we collaborated with medical experts to devise two comparative visualization techniques at different stages of the analysis process.
  First, a progressive visualization method is proposed for comparing multiple state-of-the-art causal discovery algorithms. The method can handle mixed-type datasets comprising both continuous and categorical variables and assist in the creation of a fine-tuned causal graph of a single outcome.
  Second, a comparative graph layout technique and specialized visual encodings are devised for the quick comparison of multiple causal graphs.
  In our visual analysis approach, analysts start by building individual causal graphs for each outcome variable, and then, multi-outcome causal graphs are generated and visualized with our comparative technique for analyzing differences and commonalities of these causal graphs.
  Evaluation includes quantitative measurements on benchmark datasets, a case study with a medical expert, and expert user studies with real-world health research data.
}

\keywords{Causal graph visualization and visual analysis, causal discovery, comparative visualization, visual analysis in medicine.
}

\teaser{
  \centering
  \includegraphics[width=0.95\linewidth]{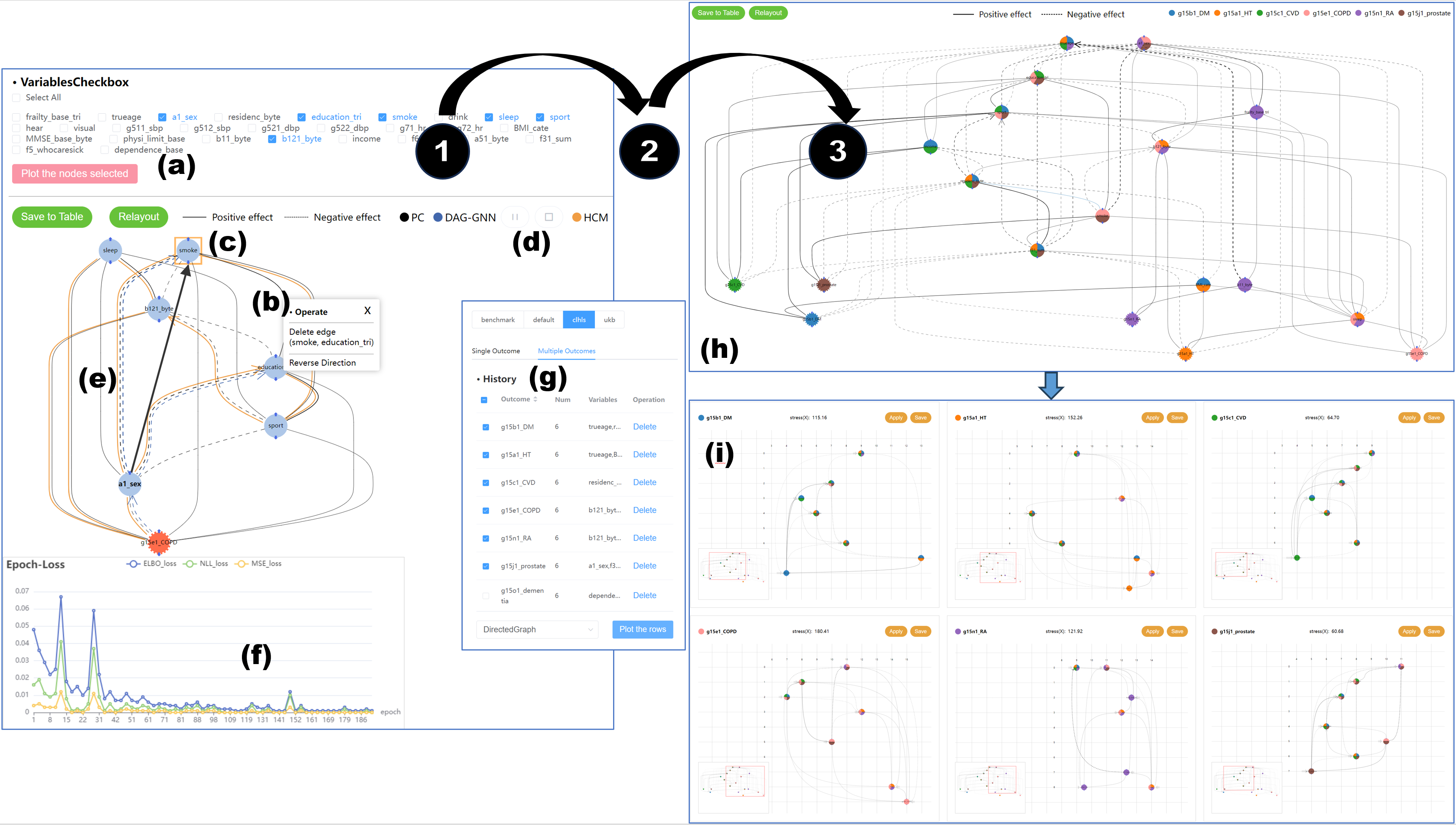}
  \caption{%
  The overall process of the visual analysis of multi-outcome causal graphs of a health research dataset~\cite{DVN2020}. Analysts use (1) the Single Graph View to analyze and edit the causal graph (a--c) of a single outcome aided
 by a set of new visualization techniques (d--f) to leverage the synergy of various causal discovery methods. The fine-tuned causal graphs of single outcomes are stored in (2) the History View (g). Next, causal graphs of interest are selected for visual comparison in (3) the Multi-outcome Graphs Comparison View (h, i), supported by our new graph layout and visual mappings (i).
  } 
  \vspace{-0.5em}
  \label{fig:interaction_process}
}

\graphicspath{{figs/}{figures/}{pictures/}{images/}{./}} 

\usepackage{tabu}                      
\usepackage{booktabs}                  
\usepackage{lipsum}                    
\usepackage{mwe}                       
\usepackage{amsmath}
\usepackage{amsfonts}
\usepackage{enumitem}
\usepackage{algorithm}
\usepackage{algpseudocode}
\usepackage{lscape}
\usepackage{mathptmx}                  
\usepackage{multirow}
\usepackage{color}

\newcommand{\indep}{\rotatebox[origin=c]{90}{$\models$}}
\begin{document}


\firstsection{Introduction}

\maketitle
Understanding causality is fundamental in human intelligence~\cite{Chou2022} and is crucial in many applications, for example, guiding practitioners designing and implementing public interventions in policy-making~\cite{Busetti2023}, assisting legal case matching in legal processes~\cite{Sun2023}, predicting how the molecular system responds to different interventions in biology~\cite{Triantafillou2017}, and assisting disease diagnosis in medicine~\cite{Richens2020}.
Causal discovery refers to the process of learning graphical structures with a causal interpretation~\cite{Zanga2022}
and has become one of the core research topics of statistics, machine learning, and artificial intelligence~\cite{Holzinger2019, Sanchez2022,Zanga2022}.
These graphical structures are directed acyclic graphs (DAGs), namely causal graphs, where vertices are data variables and directed edges denote causal relationships between variables. 

In medicine, experts also seek to discover causalities between potential risk factors and a specific disease, aiming to explore factors influencing the development of a particular disease and eventually intervene in the prognosis. 
Except for the typical case of studying the causality of a single disease (a single outcome variable), understanding the similarities and differences between different yet related causal relationships among different diseases, i.e., multiple outcome variables, is also important. For example, a study 
into the co-occurrence of more than one chronic condition in one person (multimorbidity~\cite{Harrison2021}) and the combined effect of multiple conditions in one person (comorbidity~\cite{Valderas2009}) is beneficial for proper interventions and treatments.
Therefore, it is necessary to perform analysis and comparisons for causality of multiple outcomes. 
Unfortunately, most previous visual analysis works intend to address problems with only one particular outcome.

In this paper, we study the causalities of multiple outcome variables that have not been considered in visual analytics.
Based on a requirement analysis in close collaboration with medical experts, we propose a visual analysis framework as shown in~\cref{fig:interaction_process}.
The framework consists of two stages: the discovery and interactive fine-tuning of a causal graph of a single outcome (\cref{fig:interaction_process}(1)), and the comparative analysis of causal graphs of multiple outcomes (\cref{fig:interaction_process}(3)).

Causal graphs in the first stage are calculated using three causal discovery algorithms: a classic causal discovery method that delivers a basic graph, and two state-of-the-art methods that provide analysts with more suggestions.
The resulting graphs are visualized progressively as node-link diagrams with a superposed comparative visualization for easy detection of differences.
The underlying machine learning methods can be interactively controlled, e.g., by pausing and resuming, to reduce computation time.
Users then interactively fine-tune the single outcome causal graph with reference to multiple views of multivariate data visualizations and with their domain knowledge.

Next, the user selects causal graphs of outcomes of interest for the second stage (\cref{fig:interaction_process}(g)). 
Individual graphs of single outcomes are then visualized as line-ups with a graph layout method and visual designs that facilitate comparisons (\cref{fig:interaction_process}(i)).
Our graph layout method preserves the relative position of shared nodes of supergraphs (\cref{fig:interaction_process}(h)) calculated by combining multiple single graphs while compressing nodes that may be far apart. 
We use visual designs of glyphs, the shape of nodes, and grids to assist the comparison task.

We evaluate our method with numerical measurements, a case study, and an expert user study with medical experts.
The effectiveness of the analysis and visualization of single outcome graphs is demonstrated using benchmark datasets; the multi-outcome comparison design is evaluated with stress measurements.
The overall visual analysis method is evaluated through a case study of the analysis of diseases that are usually seen in multimorbidity/comorbidity using two widely accepted health research datasets.
Feedback from expert users indicates that our method is useful for supporting causal reasoning of single and multiple outcomes and provides insights with the comparison of multi-outcome causal graphs that are impossible with existing methods.

Our major contributions are as follows.
\begin{itemize}
    \item Two-stage visual analysis of multi-outcome causal graphs.
    \item Progressive visualization and interactive controls of multiple causal discovery techniques including modern continuous causal discovery methods.
    \item Graph layout and visual designs that support effective comparison of multiple DAGs.
    \item A case study using real-world health research datasets with a medical expert.
\end{itemize}
Please note that our method aims to address a domain problem in healthcare but is also general enough to aid causal analysis of single and multiple outcomes. 
The source code of our visual analysis system is publicly available at:~\url{https://github.com/inco-yjl/multi_outcome}.

\section{Related Work}
Our work is related to three areas of research: causal discovery, visualization and visual analysis of causality, and comparative visualization.

\subsection{Causal Discovery}
Causal reasoning is a challenging and active topic in machine learning, and details of this topic are covered in textbooks~\cite{Pearl2009, Peters2017}. 
Causal discovery is the process of finding causal graphs of variables in a given dataset~\cite{Zanga2022}.
The state-of-the-art of causal discovery is summarized in surveys~\cite{Vowels2022,Zanga2022}.
The methods can be categorized into constraint-based, score-based, structural asymmetries-based, and those exploiting various forms of intervention.
In terms of how the problems are solved, two classes of techniques are available: traditional combinatoric approaches and modern continuous methods. 

Combinatoric methods search for DAGs satisfying criteria within the aforementioned four categories.
Popular combinatoric methods include the Peter-Clark (PC) algorithm~\cite{Spirtes1993}, Greedy Equivalence Search (GES)~\cite{Chickering2003}, satisfiability (SAT) solver~\cite{Hyttinen2013}, and LinGAM~\cite{Shimizu2006}.

In contrast, continuous methods rely on continuous optimizations that are typically used in neural networks.
NO TEARS (Non-combinatoric Optimization via Trace Exponential Augmented Lagrangian Structure learning) is considered the first to formulate the combinatorial graph search method as a continuous optimization problem~\cite{Zheng2018}.
A further improvement of the method that is faster and more accurate than NO TEARS is available~\cite{Bello2024}.
DAG-GNN~\cite{Yu2019} is a graph neural network approach to solve the continuous optimization problem. 
Methods focused on categorical data are also available with continuous optimizations using interventional or reinforcement learning~\cite{Ke2020, Brouillard2020}.

One issue with causal discovery methods is that they typically assume that all variables have the same data type---either categorical or continuous. 
In real-world datasets, however, data types are mixed, for example, typical survey data in medicine includes continuous biological measures like body height, weight, and blood pressure, and binary/categorical attributes like gender and stages of diseases.
Few, dedicated methods are available for mixed-type datasets~\cite{Andrews2018, Andrews2019, Li2022,Cai2022}, while some others, for example, DAG-GNN~\cite{Yu2019}, can handle such data inherently.
A latent-PC algorithm is available for multi-dimensional mixed-type data that is typical in medical research~\cite{Cai2022}.

\subsection{Visualization of Causal Graphs and Visual Analysis of Causal Reasoning}
The visualization of causal graphs is studied in several works~\cite{Bae2017, Xie2021, Vo2020, Cottam2021}.
Different visual mappings of the graph, e.g., adjacent matrix and node-link diagram are compared in a user study~\cite{Vo2020}.
For node-link diagrams, different graph layouts are also compared with user studies~\cite{Bae2017, Vo2020}. 
Layered graph layouts are used for showing large causal graphs~\cite{Xie2021} or comparing different causal discovery results~\cite{Cottam2021}.

Visual analysis methods for causal reasoning are available to aid the understanding of causalities and support interactive decision-making.
An interactive causal reasoning interface allows visually editing and verifying causal relationships with multiple linked views of the causal graph, and statistical parameters~\cite{Wang2016}.
A more comprehensive extension of the method is available with the support of multidimensional data visualizations, a visual design of the causal graph with layered graph layout, and the capability of handling mixed-type data~\cite{Wang2017}.
An exploratory causal analysis approach introduces the uncertainty-aware visualization of causal graphs and provides
interactive means for exploring, validating, and applying causal relations for decision-making~\cite{Xie2021}.

Some works focus on the understanding of causal inferences and counterfactuals~\cite{Lewis1973}.
A multi-view visualization tool is available to support iterative causal inference~\cite{Guo2023}.
Counterfactuals are used to improve the visualization of relationships between variables~\cite{Kaul2022} and causal inferences~\cite{Borland2024}.
Visual analytics of time-dependent causality can be applied to event sequences with a user-modifiable Granger causality model~\cite{Jin2021}.
Causal hypotheses of time-dependent events can be formulated and tested with time delay in a visual analysis method~\cite{Wang2023}. 

Visual causal analysis is also used to address domain-specific problems.
The interpretation of algorithmic decision-making models is supported by a visual analysis method that exploits the explanatory ability of causal models~\cite{Hoque2022}.
Algorithm bias is tackled with a causality-based visual analysis method that allows users to audit the data and generate debiased data~\cite{Ghai2023}.
Questionnaire responses are analyzed using causal reasoning for question combinations with association mining and causal sub-graphs of each question combination are visualized and explored in a visual analysis system~\cite{Li2024}.
Spatial-temporal urban data is detected and analyzed using the Granger causality test in multiple linked views~\cite{Deng2022}.

However, none of the above works utilizes state-of-the-art continuous optimization methods nor studies the comparison of causal graphs of multiple outcome variables in a dataset.
Our method differs from these works in two important aspects: we use state-of-the-art causal discovery methods and address the problem of causal reasoning and visual comparison of multiple outcome variables that are important for public health. 

\subsection{Comparative Visualization}
It is a common yet challenging task to design methods supporting comparison in visualization~\cite{Gleicher2018}. 
Four considerations are presented for comparative visualization designing and evaluation~\cite{Gleicher2018}. 
A general taxonomy of visual comparisons classifies visual designs into three basic categories---juxtaposition, superposition, and explicit encodings~\cite{Gleicher2011}.
These categories should be combined for effective comparative visualization tools and tasks. 
A qualitative user study shows that interactive techniques are important to visual comparison tasks~\cite{Tominski2012}. 
A combined encoding of colors and positions enables analysts to easily view differences and changes in the comparison of task-driven topic models~\cite{Alexander2016}.

Research efforts have been made for the visual comparison of graphs.
Visual comparison of two graphs is achieved by visualizing their similarities and differences in a merged graph~\cite{Andrews2009}.
Interactive graph matching~\cite{Hascoet2012} addresses the visual comparison of graphs and clustered graphs.
Weighted graph comparison is explored in brain connectivity analysis~\cite{Alper2013}, where the effectiveness of node-link diagrams and adjacency matrices is compared. 
A map-based abstraction of node-aligned graphs on a triangle mesh facilitates a clutter-reduced comparison of undirected graphs~\cite{Jin2021a}.  
Layout optimization, graph embedding, and clustering and grouping algorithms are used for the comparison of two temporal graph datasets~\cite{Zhang2022}.

Similar to these works, one of our goals is to visually compare graphs. 
However, existing works mainly address the comparison of two undirected graphs, whereas our method is designed to compare more than two directed graphs.

\section{Requirement Analysis and Method Overview} 
Our work is a long-term collaboration with medical experts, which started in 2022. 
The work was initially motivated by the needs of one of our collaborators who studies the coexistence of multiple clinical conditions or diseases, such as multimorbidity or comorbidity, which are important topics in clinical medicine and public health~\cite{Tazzeo2021,Hanlon2018}.
Mechanisms underlying multimorbidity or comorbidity are complex, they may share some direct causation, and associated risk factors like aging and socioeconomic deprivation~\cite{Olson2018, Valderas2009, Skou2022}, but the same factors may play different roles in different diseases. 
The expert hopes to analyze the influencing factors of different diseases and compare those diverse yet correlated diseases or clinical outcomes aiming to gain insights into the development of diseases and improve prognosis.

Traditionally, medical experts use statistical methods such as multiple logistic regression to explore factors associated with a disease.
However, it is well-known that ``correlation does not necessarily imply causation''~\cite{Liang2021}, and, therefore, we considered using causal analysis to address this domain problem of analyzing multiple outcomes. 

\begin{figure}[tb]
  \centering
  \includegraphics[width=0.93\linewidth]{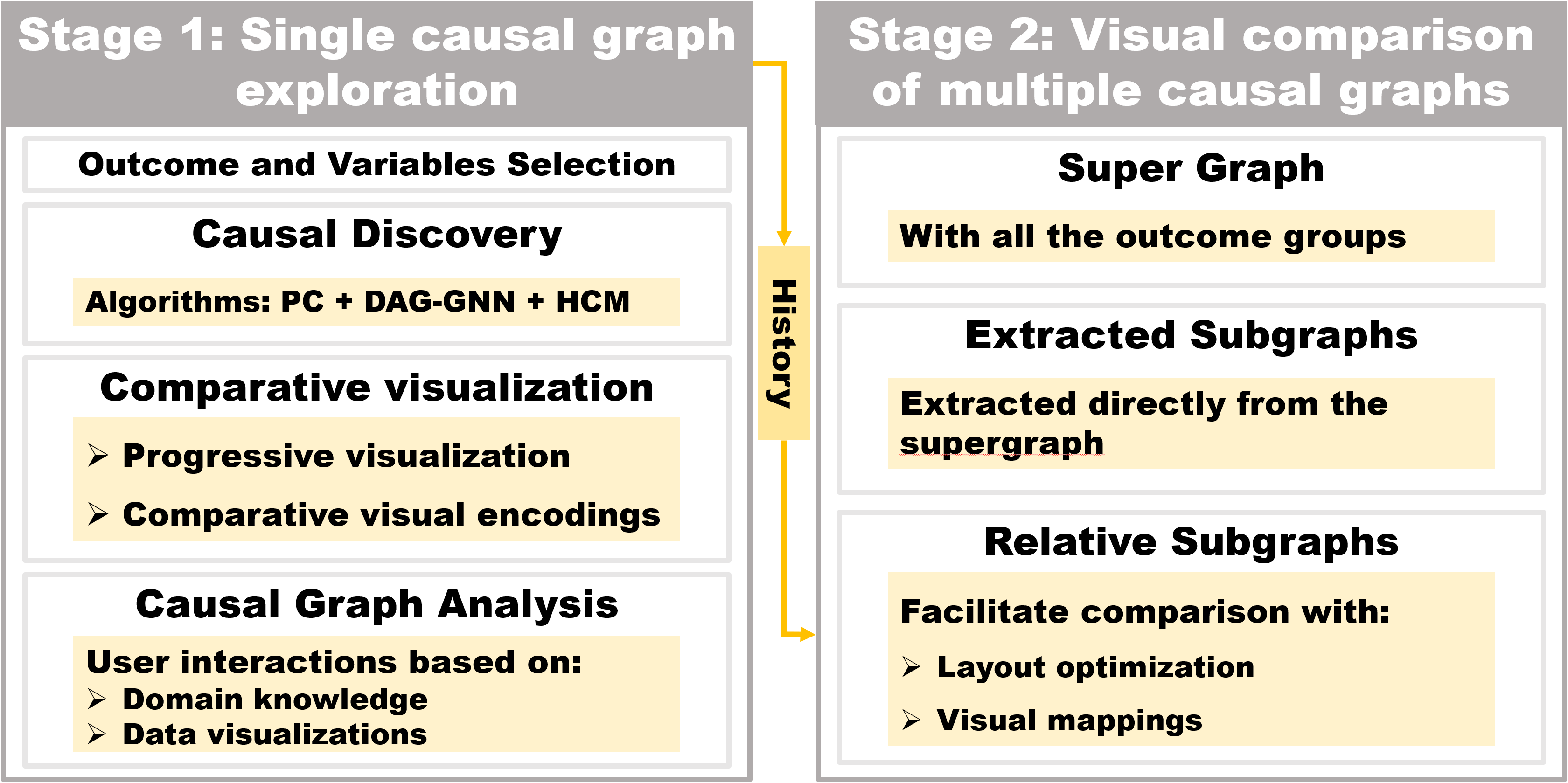}
  \caption{%
  The workflow of our visual analysis method.
  }
    \vspace{-0.5em}
  \label{fig: workflow}
\end{figure}

Following the design study methodology~\cite{Sedlmair2012} and the abstraction according to a nested model for visualization design~\cite{Munzner2009}, we hosted one in-person meeting in the beginning (July 2022) and several follow-up online discussions (roughly every 2--3 months depending on the progress) with three clinical medicine and public health experts from the university that the first/corresponding author affiliates to (one of them is a co-author who is now at a different university).
First, we abstracted the domain problem into two data analysis stages as shown in~\cref{fig: workflow} that explains our workflow.
\begin{enumerate}[label=\textbf{S\arabic*}]
    \item \textbf{Single causal graph exploration and analysis.} Experts hope to analyze the causality between a specific clinical outcome and its potential influencing factors efficiently from a large dataset. The causal graph should be modified interactively by the user and the visualization should provide good interpretability for the underlying causal learning techniques.
    \item \textbf{Multiple causal graph comparison and analysis.} In this stage, experts hope to select interested clinical outcomes and compare causal graphs of the corresponding outcomes to guide interventions and treatments for different diseases.
\end{enumerate}

We initially started with a design study on this specific healthcare problem and realized that aspects of this problem are general in causality visualization and visual analysis.
For example, our problem involves the layout for causal graphs~\cite{Bae2017,Vo2020,Xie2021}, interactive editing of causal graphs~\cite{Wang2016,Wang2017,Deng2022}, comparison of causal learning methods~\cite{Cottam2021}, the interpretation of causal graphs~\cite{Wang2016,Wang2017,Hoque2022,Deng2022}, and causal subgraphs~\cite{Li2024}.
Therefore, combining feedback from domain experts with design principles from previous causal-analysis-related visualization works, we further specify the following design requirements.
\begin{enumerate}[label=\textbf{R\arabic*}]
    \item \textbf{Support the interpretation of causal graphs.}
    The causal graphs should be interpreted by domain experts with ease.
    Clearly visualize causal graphs with a reasonable graph layout and provide effective visual encoding of the information of nodes and edges.
    \item \textbf{Visual comparison of causal graphs computed by various causal discovery methods.}
   None of the existing causal discovery methods always yields good results in all cases.
    Visualize results of multiple causal discovery methods with comparative visualization as a reference to analysts for constructing a reasonable causal graph.
    
    \item \textbf{Improve the efficiency and interpretability of causal discovery.} 
     Long run time and poor interpretability are typical issues of causal discovery methods, e.g., modern deep-learning methods are too slow to converge for an interactive visual analysis system.
     Visualize intermediate results and diagnostic measures of the model quality to guide a user-controllable causal discovery process balancing the speed and quality. 
     \item \textbf{Assist the selection of variables for analysis.} 
     Datasets with multiple clinical outcomes typically contain a large number of variables. 
     Including all variables would result in an overwhelmingly complex causal graph that prohibits understanding. 
    Provide statistical and visualization methods combined with interactive approaches to help analysts decide possible related variables.

    \item \textbf{Incorporate human knowledge into causal analysis and causal graph editing.} To obtain real causal relationships with causal discovery algorithms is unlikely. 
    Adding expert domain knowledge to the exploration and modification of causal graphs through interactions enhances the reliability of the results.
    Basic interactions should include adding and deleting nodes and edges, and inverting edges.
    \item \textbf{Allow intuitive comparison of causal relationships for different outcomes.} Different diseases may share the same influencing factors, and distinguishing the shared and different influencing factors can help medical experts rationally judge and intervene in specific influencing factors, and improve prognosis.
\end{enumerate}

\section{Visualization of Single Causal Graph with Various Causal Discovery Methods}
\label{sec:singleGraphStage}
In this section, we introduce the selected causal discovery methods, elaborate on the visualization of single causal graphs with the rationale of design choices, and explain the causal effect computation. 

\subsection{Causal Discovery}
\label{sec:causalGraphMethod}
There are various causal discovery methods~\cite{Vowels2022}, but none is currently available as an all-around winner.
Therefore, we aim to use recent causal discovery methods that inherently handle mixed-type data with categorical and continuous variables and could provide complementing results so that the user can combine them to get a good causal graph.
Three causal discovery methods are adapted into our system: the classic Peter-Clark (PC)~\cite{Spirtes1993} algorithm, which is known to provide a good causality skeleton, the DAG-GNN method~\cite{Yu2019} as a representative modern continuous optimization method that can handle mixed-type data, and the hybrid causal learning method (HCM)~\cite{Li2022}, specifically designed for mixed-type data. 
More details of the methods are covered in the supplemental material.

\subsubsection*{(1) PC}
The PC~\cite{Spirtes1993} algorithm is a well-known and frequently used causal discovery method that serves as the backbone approach.
The method starts with a fully connected undirected graph and identifies a skeleton by evaluating the conditional independence of each variable pair.
Next, the edges are oriented by identifying colliders/v-structures. 
The algorithm then orients additional edges in partially directed paths that would otherwise form more v-structures.

\subsubsection*{(2) DAG-GNN}
The original combinatorial optimization problem of causal discovery can be transformed into a continuous optimization problem~\cite{Zheng2018}.
DAG-GNN~\cite{Yu2019} uses a generative graph neural network based on variational autoencoders with neural network functions and evidence lower bound to solve the problem.
It extends the NO TEARS~\cite{Zheng2018} method for nonlinear structural equation models and supports both discrete and continuous variables.

\subsubsection*{(3) HCM}
HCM is a recent advancement for mixed-type data.
It addresses the mixed-type data problem by formulating a mixed structure equation model~\cite{Li2022} that treats continuous and categorical variables separately.
The model is learned with skeleton learning using the PC-stable algorithm~\cite{Colombo2014} with a new mixed-type conditional independence test; then the method finds causal graph directions via greedy search with a mixed information criterion; and the graph is further pruned using the conditional independence test.

\subsection{Evaluation with Benchmark Datasets}
\begin{figure}[htb]
    \centering
     \includegraphics[width=0.8\linewidth, trim={0 0.1cm 0 0},clip]{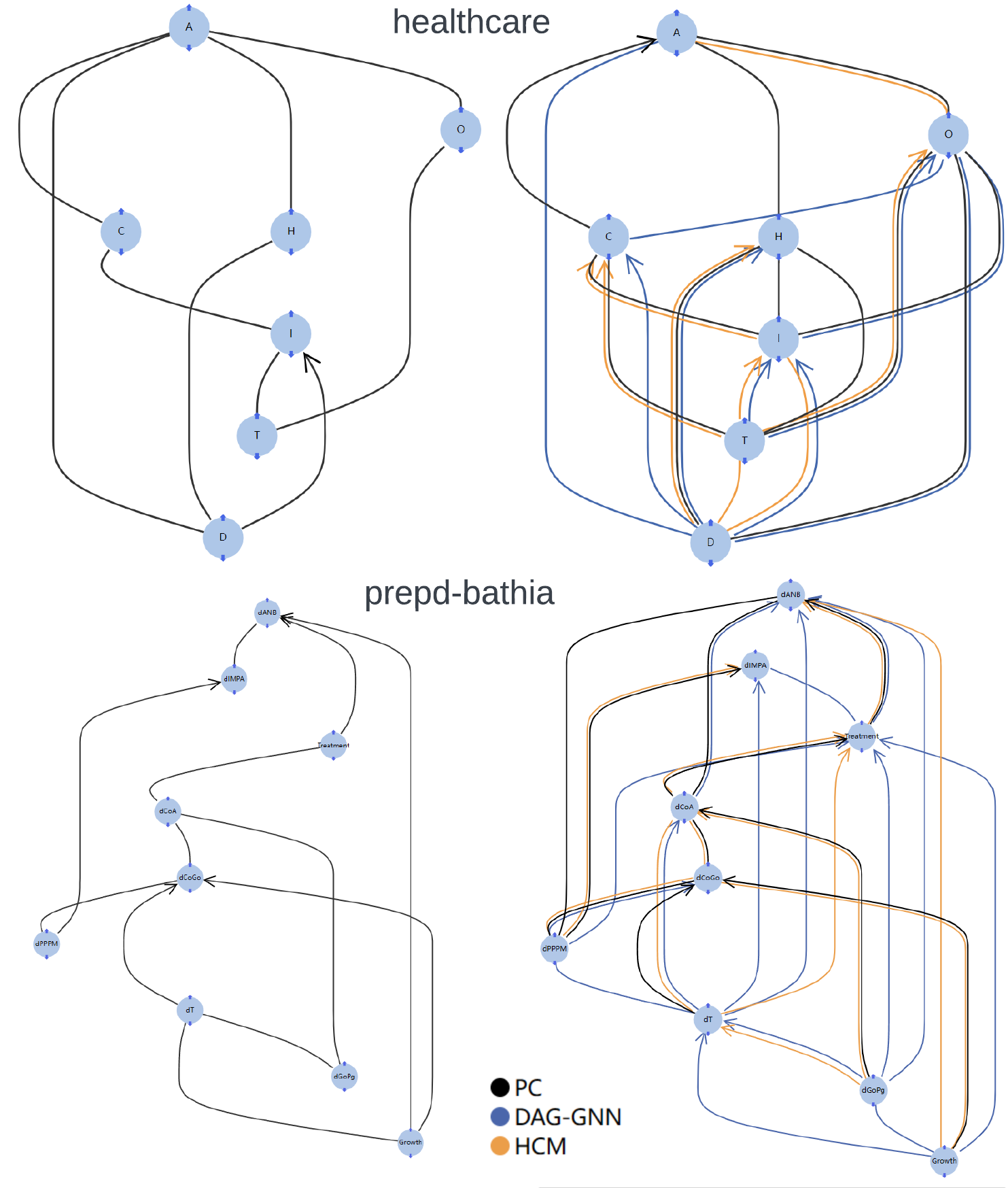}
    \caption{Visualization of the true causal graphs (left column) and results of the three causal discovery algorithms (right column).}
    \vspace{-0.5em}
    \label{fig:algCompareBenchmark}
\end{figure}

The effectiveness of using multiple causal discovery methods is demonstrated using two mixed-typed benchmark datasets: healthcare\footnote{\url{https://www.bnlearn.com/bnrepository/clgaussian-small.html\#healthcare}} and prepd-bathia\footnote{\url{https://www.bnlearn.com/research/scirep17/}}.
Results of the three algorithms are visualized in \cref{fig:algCompareBenchmark}(right), with our visualization method (\cref{sec:compInsituVis}).

\begin{figure*}[tb]
        \centering
    \includegraphics[width=0.9\linewidth]{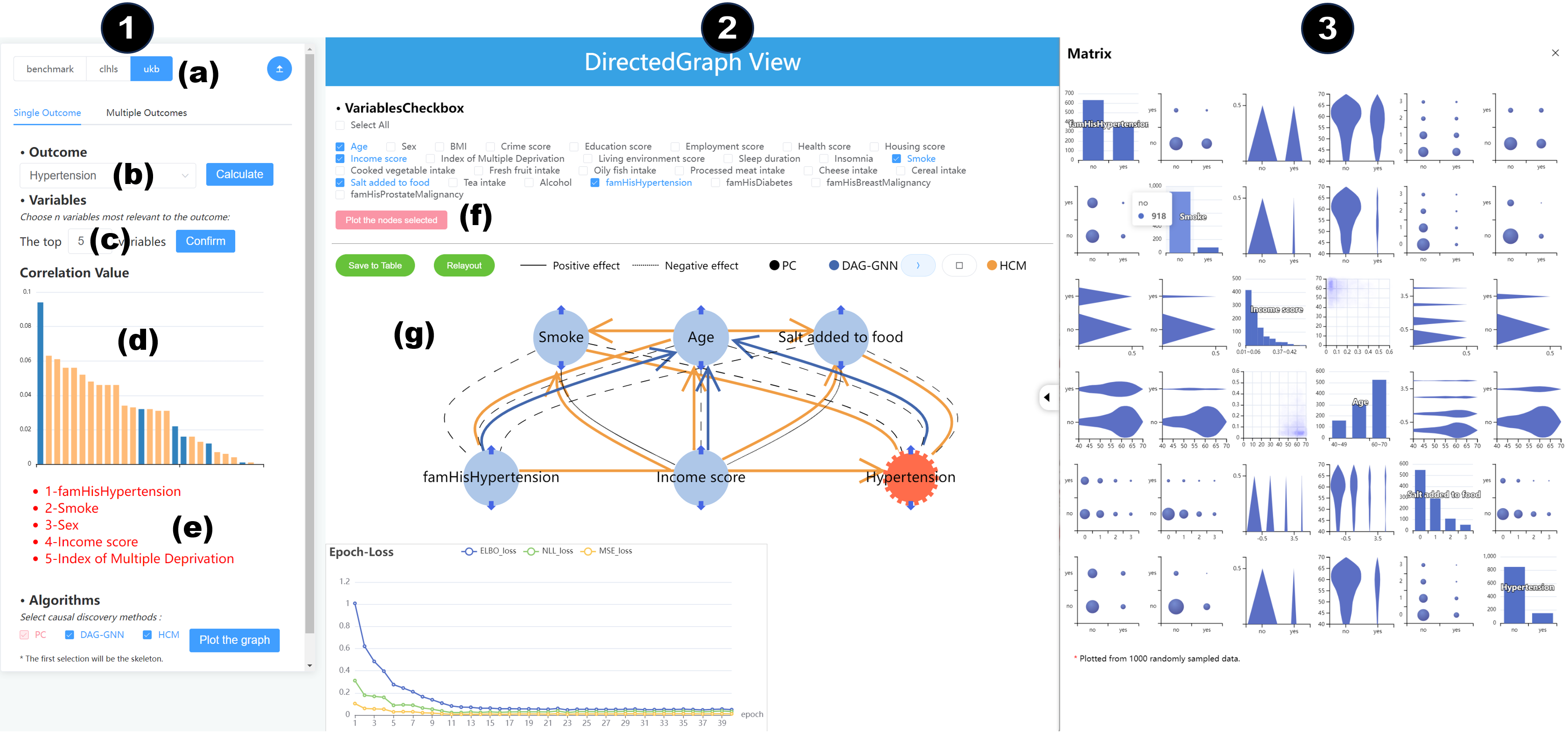}
    \caption{The user interface of our visual analysis system for single outcome graphs consists of three views: 
    (1) Dataset and Variables Selection View allowing for dataset (a) and variables selection (b, c) and correlation analysis (d, e), (2) Single Directed Graph View allowing for graph editing (f, g), and (3) Variables Matrix View providing insight for graph editing. 
    }
   \vspace{-0.5em}
    \label{fig:sgUI}
\end{figure*}

The combined results of all three algorithms yield an (undirected) accuracy of 100\% and 92\% compared to the true graph (\cref{fig:algCompareBenchmark} (left)), respectively, at the cost of high false positive rates.
More numerical measurements are summarized in the supplemental material.
With our visualization, false positives are not a severe issue compared to a lower accuracy because removing a wrong edge is easier than discovering a new true edge---the presence of a wrong edge is easily observable given the domain knowledge of the analyst.

Moreover, with our visualization, patterns can be found that are otherwise difficult with numerical measures only.
For example, the coexistence of edges discovered by different methods suggests that the possibility of a true edge is high.
In both benchmark datasets, edges shared by all three methods are all true edges.
Therefore, the combination of these algorithms provides users with important additional information for creating a better causal graph than using any of them alone.

\subsection{Comparative Progressive Visualization of Causal Discovery Algorithms}
\label{sec:compInsituVis}
Our visualization method for single-outcome causal graph addresses the following three issues: (1) it visually encodes information of a causal graph (\textbf{R1}), (2) supports the visual comparison of results generated by the algorithms in~\cref{sec:causalGraphMethod} (\textbf{R2}), and (3) visualizes the progress of the algorithms with user-controllable pausing and resuming (\textbf{R3}).

\subsubsection{Visual Encodings of Causal Graphs}
\label{sec:vis_causal_encode}

An efficient layered method\footnote{\url{https://github.com/dagrejs/dagre/wiki}} that takes advantage of several graph drawing works~\cite{Gansner1993, Jünger2002, Barth2002, Brandes2002} is used for the layout of the causal graph.
In this layout, nodes are arranged top-down with decreasing out-degrees, and the cause-outcome relationship is inherently encoded by the vertical locations of two variables, i.e., the direction of causality naturally is from the upper node to the lower node of the two nodes connected by an edge.  
However, results of causal discovery methods sometimes include reversed causalities with the layered layout, i.e., cause nodes are below effect nodes.
Such potentially self-contradictory cases require user attention. 
Arrows are commonly used to encode directional properties,
so we choose to highlight the reversed causality using an edge with an arrow heading from the cause to the effect and keep other edges with top-down causal directions undirected to reduce clutter.

Color and shape---known as effective visual channels for categorical variables~\cite{Mackinlay1986}---are used to distinguish outcome and other nodes.
Edge thickness, a frequently used channel for weight encoding in a weighted node-link diagram~\cite{Alper2013}, shows the magnitude of the causal effect (\cref{sec:causalInfer}). 
Note that the causal effect is signed, and we use the edge style to encode the sign: solid lines for positive and dashed lines for negative effect values, respectively. 
Logarithmic transformation is applied to adapt the large range of the effect value.

Example visualizations are shown in~\cref{fig:interaction_process}(1), \cref{fig:sgUI}(g), and \cref{fig:case_hypertension}(4).
The node encoding uses jagged circles in orange for the outcome variable, and regular circles in blue for other variables, and the name of each variable is drawn at the center of the circle.
The two small widgets (in blue) attached to the top and bottom of a node are used for adding new edges.

\subsubsection{Comparative Visualization of Causal Discovery Results}
An appropriate visual comparison design for complementing causal discovery results could aid the editing of the final causal graph. 
We consider the comparison strategies~\cite{Gleicher2011} for our case that causal graphs have the same nodes but different edges.
For this case, juxtaposition is slow and error-prone for comparison~\cite{Alper2013}, and explicit encodings computing a number of pairwise relationships can be confusing and cluttered.
In contrast, superposition that directly visualizes the presence/absence of an edge is an effective and space-efficient solution.

Our comparative visualization uses the superposition of the PC, DAG-GNN, and HCM results.
Color coding is used for differentiation: black for PC, blue for DAG-GNN, and orange for HCM.
Jittering is used to show the three results simultaneously without occlusion: the PC result is drawn at the center between nodes, the DAG-GNN result is offset to the right, and the HCM result is to the left.
By default, all three algorithms are shown together. 
Each result can be highlighted by assigning it a high opacity while reducing the opacity of others with the legend shown on the top right of the view.
Moreover, for showing a specific result or comparing any of the two, the rendering of each result can be conveniently switched on and off.

\subsubsection{Progressive Visualization}
The progressive visualization addresses the conflict between the long running time of causal discovery methods and the interactivity of visual analysis.
For example, the DAG-GNN method by default runs for a large number of epochs, where each epoch takes a few seconds, making a naive use of it impractical.
Instead, we progressively draw intermediate results at each epoch by updating the graph edges within the current layout, so that the running of the causal discovery methods does not prohibit the visual analysis.

Diagnostic measures of DAG-GNN are visualized as line charts in an epoch-loss view shown at the bottom of~\cref{fig:interaction_process}(f), \cref{fig:sgUI}(g), and \cref{fig:case_hypertension}(4).
Specifically, the measurements---ELBO (Evidence Lower Bound) loss (blue), NLL (Negative Log-Likelihood) loss (green), and MSE (Mean Squared Error) loss (orange)---provide interpretability of the method and evidence for users to determine when to pause or stop the algorithm.
Users can pause the process at any time to examine the current result and stop or restart the process depending on whether the result is satisfactory or not.
Typically, the losses drop quickly after a few dozen epochs and become stable, and similar is true for the associated causal graph result.
Therefore, our visualization provides more insights into the DAG-GNN method than what was previously possible.

\subsubsection{Multidimensional Visualization of Mixed-Typed Variables}
To gain more insights into the variables of the causal graph, a multidimensional visualization is available in our method to show the distributions of variables of mixed types.
As illustrated in~\cref{fig:sgUI}(3), the visualization resembles the scatterplot matrix (SPLOM) showing distributions of every pair of variables.
Unlike a SPLOM, we apply different visual encodings according to the type of variables according to the conventions of the domain experts.
This design is suitable for our datasets (random sampling) as the visualizations are readable since the data are rather sparse and the encodings do not generate clutter.

For categorical vs. categorical, bubble plots that encode the number of samples with area, are used due to the familiarity of the domain experts.
Violin plots---effective for comparing continuous distributions under different categories---are used for categorical vs.\ continuous.
Scatterplots are used for continuous vs.\ continuous as they can clearly show distributions of two continuous variables.
We use the whole matrix for the visualization as it is found more understandable and convenient for searching by domain experts than the triangular version.

\begin{figure*}[htb]
    \centering
    \includegraphics[width=0.9\linewidth]{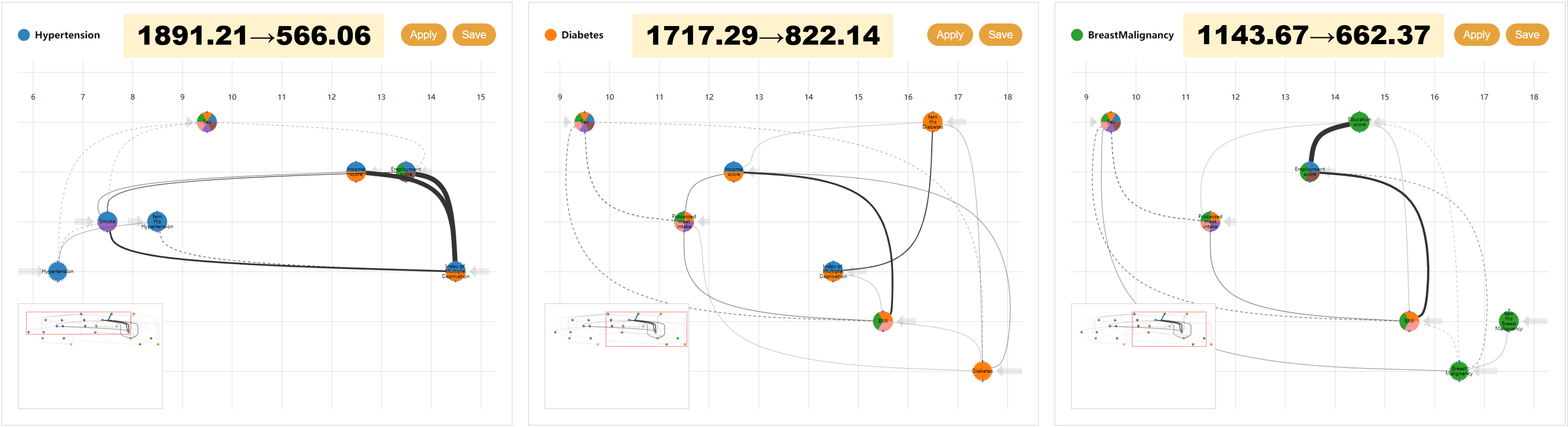}

    \caption{Visualization of multi-outcome causal graphs with our new layout and visual mappings. Numbers on top of each subgraph are the horizontal stress of direct extraction (left of the arrow) and our method (right of the arrow), respectively.}
        \vspace{-0.5em}
    \label{fig:causal_comparison}
\end{figure*}

With our comparative progressive visualization of the causal graph and the multidimensional visualization of variables involved, analysts can then interactively refine the causal graph of a single outcome with their domain knowledge.

\subsection{Causal Effect Computation}
\label{sec:causalInfer}
The causal effect of a variable X on Y is defined as the expected change in Y due to a change in X. 
In our case, we apply the backdoor identification algorithm~\cite{Pearl2010} and linear regression estimation to estimate the causal effect of each edge.

\section{Visualization of Multi-outcome Causal Graphs}
\label{sec:multigraphsVis}
Once the individual causal graphs for individual outcomes have been analyzed and refined, our method supports the visual comparison of multi-outcome causal graphs composed by these individual graphs.
In this section, we introduce the graph layout method for comparing multiple causal graphs and explain the associated visual designs facilitating comparisons.

\subsection{Graph Layouts for Comparison}
\label{sec:layout_explore}

We first experimented with the direct comparison of subgraphs obtained using the aforementioned layout algorithm individually.
We found that each subgraph reaches its optimal layout, but comparing these subgraphs is difficult.
This is because the information of other subgraphs is missing during the layout computation and the subgraphs live in their own spaces that are not comparable. 
Therefore, it is necessary to include global information of all subgraphs for comparison.
To this end, we propose two layouts for comparison, namely, the supergraph (for 2 to 3 subgraphs) and the comparable graph layout (for more subgraphs).

\subsubsection{Supergraph of All Outcomes}

We combine all single graphs of interest 
as one large causal graph, namely, the supergraph $G_s$. 
The supergraph is laid out using the aforementioned layered method and an example is shown in~\cref{fig:interaction_process}(h). 
According to the convention of the layout method, the vertical indices are called \emph{ranks}, and the horizontal indices at each rank are \emph{orders}.

The supergraph can show the similarities and differences between subgraphs because of the global image space and anchored common nodes.
The supergraph properly supports comparing up to two or three outcomes but does not scale well for more outcomes perceptually.
As shown in~\cref{fig:interaction_process}(h), comparing causal graphs of each outcome is difficult if not impossible with the supergraph of six outcomes.
For more than three outcomes: in cases with few shared nodes, the supergraph becomes too wide; for many shared nodes, the association between nodes and subgraphs is difficult to perceive (shared nodes have several colors in our case).
Instead, it is advantageous to compare individual graphs as line-ups (\cref{fig:interaction_process}(i)) with comparable graph layouts.

A typical approach for subgraph line-ups would directly extract subgraphs of individual outcomes while preserving the locations of nodes and edges in the supergraph.
This is the strategy used in the visual analysis of multiple biological interaction graphs~\cite{Barsky2008}.
However, subgraphs of individual outcomes extracted directly from the supergraph are rather sparse, as shown in~\cref{fig:causal_comparison}.

\subsubsection{Comparable Layout of Multiple Causal Graphs}
\label{sec:layout_optimize}
Based on our experiments with various layout strategies, we decided to optimize the layout of subgraphs extracted from the supergraph to facilitate the comparison of more than two graphs. 
A key feature considered during the design of the layout algorithm is to preserve the relative node positions of the supergraph to make them comparable.
Therefore, the objectives of the comparable layout are abstracted as: 1)~preserving the relative node positions, 2) anchoring the common nodes, and 3) making the graphs compact.

The according constrained optimization problem reads for a given subgraph $G_t$ of nodes $V$ with input coordinates $\mathbf{x}$ in the supergraph, output positions $\mathbf{\xi}$, the set of indices $Q$ of all common nodes $V_Q$ shared by any subgraphs, and the index set $C_t$ of unique nodes of $G_t$:
\begin{align}
    \min_{V\in G_t}& \sum_{i,j\in Q}w_i(||\mathbf{x}_i - \mathbf{x}_j|| - ||\mathbf{\xi}_i - \mathbf{\xi}_j||)^2 +\nonumber\\
    & \sum_{k,l\in C_t}w_l(||\mathbf{x}_l - \mathbf{x}_k|| - ||\mathbf{\xi}_l - \mathbf{\xi}_k||)^2  \label{eqn:layoutGoal}\\
    \text{subject to: } & V \; \text{do not violate ranks and orders} \nonumber\\
    & V \in V_Q\; \text{are anchored} \;.    \nonumber
\end{align}

Since the supergraph already has the optimized layout~\cite{Brandes2002}, the problem of~\cref{eqn:layoutGoal} can be approximated with a fast greedy algorithm.
Note that the rank (the layer of nodes) indicates the out-degrees, and should be clearly separated and aligned between subgraphs for comparison.
We decided to leave it intact after experimentation.
Therefore, we just need to reduce the horizontal space by compressing the horizontal coordinates of nodes while maintaining the relative node positions.

Our horizontal compression algorithm first identifies common nodes shared by two or more subgraphs and compresses them horizontally to create anchored nodes $V_Q$. 
Then, for each subgraph $G_t$, unique nodes are relocated by compressing their horizontal distances to anchored nodes $V_Q$.
Finally, the compressed subgraph is achieved by merging compressed unique nodes and the anchored nodes of $G_t$.
The detailed algorithm can be found in the supplemental material.

Examples of subgraphs with the horizontal compression layout (namely, relative subgraphs) can be seen in~\cref{fig:causal_comparison}.
Compared to subgraphs extracted directly from the supergraph, our new layout provides better usage of screen space by removing much of the empty space while the relative positions are preserved.
Moreover, we numerically evaluate the effectiveness of our method by computing the horizontal stress (i.e., stress(X)) of each subgraph~\cite{Gansner2004}.
A smaller stress number indicates a better approximation of the graph's theoretical distance.
As shown on top of each subgraph, our layout method yields smaller stress(X) numbers compared to directly extracted subgraphs.
This suggests that our layout leads to edge distances closer to the graph's theoretical distances than directly extracted graphs.

\subsection{Visual Mappings}
\begin{figure}[tb]
    \centering
    \includegraphics[width=0.78\linewidth]{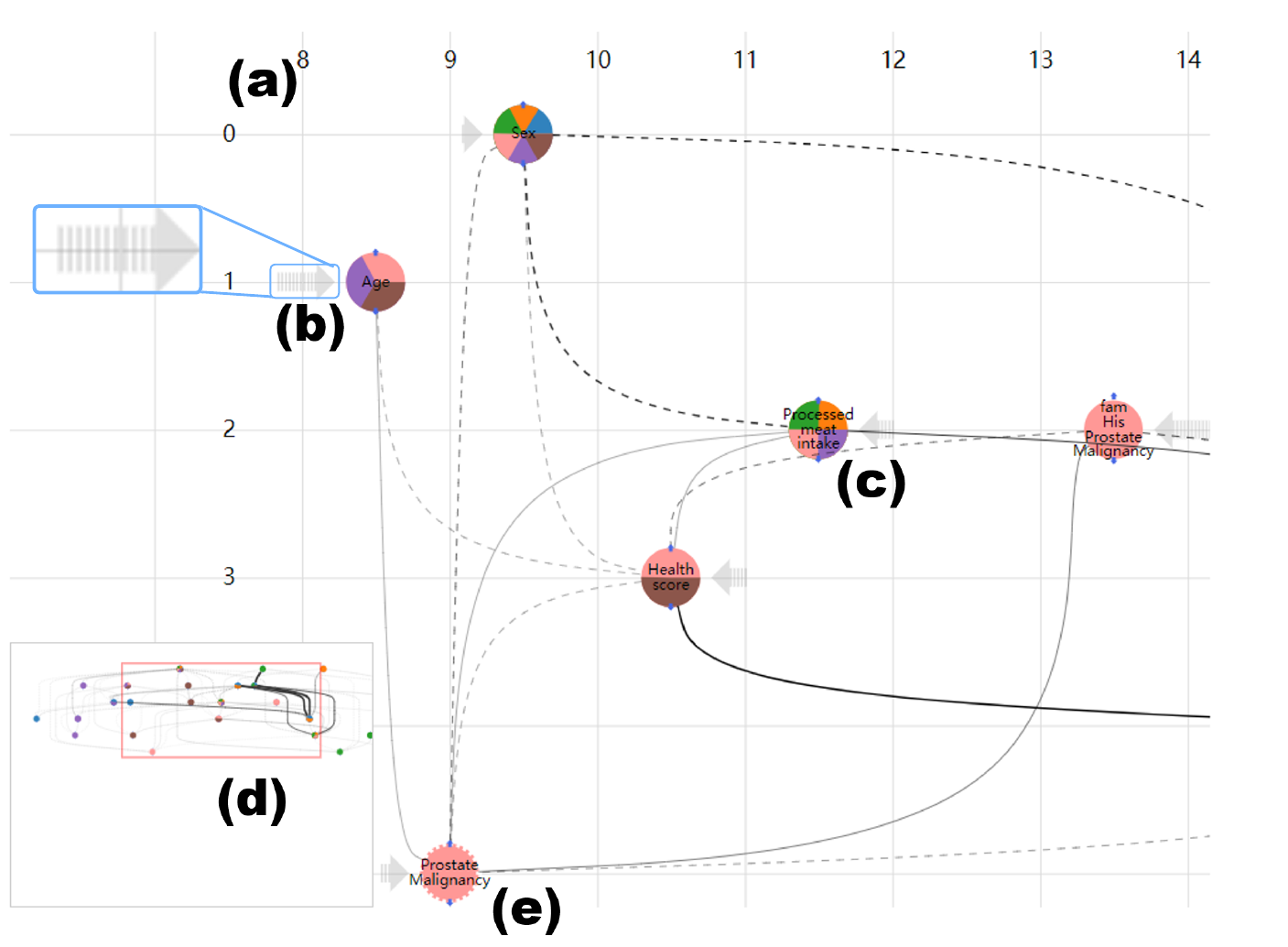}
    \caption{Visual mappings for comparison of multi-outcome causal graphs.}
    \vspace{-0.5em}
    \label{fig:multi_encoding}
\end{figure}

The node encoding (jagged and regular disks) of the single graph is extended for the comparison of multiple causal graphs.
Unlike the single graph, node color here encodes different outcome graphs.
A node shared by multiple graphs has multiple colors filling the disk area evenly divided by the number of shared graphs.
For example, some nodes shown in~\cref{fig:multi_encoding} are shared by various numbers of graphs.
The visual encoding on edges (of effect values and causal directions) is the same as in~\cref{sec:vis_causal_encode}. Rather than all three causal discovery results, only the final user-edited causal graph is shown.

With our layout method, non-anchored nodes at different original positions may land on similar or the same locations in subgraphs after the compression. 
Therefore, we augment the compressed graphs with reference grids and glyphs for accurate comparisons.
Reference grids with coordinate labels can aid the comparison of the global positions of nodes. 
The grids can be zoomed in and dragged for better comparison. 
Grids with coordinates provide quantitative information for comparison but require the shift of attention from nodes to the grid, which may potentially interfere with the mental map. 
Therefore, we design arrow glyphs (\cref{fig:multi_encoding}(b, c)) to directly encode the extent and directions of compression, i.e., left or right, on each node. 
The number of bars on the arrow glyph (the inset of~\cref{fig:multi_encoding}) encodes the moved distances relative to the position in the supergraph. 
The arrow on the left (\cref{fig:multi_encoding}(b)) of a node indicates that the node has moved to the right relative to its original position after graph compression, and vice versa for arrows on the right (\cref{fig:multi_encoding}(c)). 

To further indicate the global position of the subgraph, a thumbnail (\cref{fig:multi_encoding}(d)) is provided at the bottom left of each subgraph.
The thumbnail view shows the bounds of the subgraph (as a red box) within the supergraph for a quick lookup.

\section{Visual Analysis of Multi-outcome Causal Graphs}
\label{sec:visualAnalysis}
The visual analysis of multi-outcome causal graphs is divided into two phases: single causal graph exploration and multi-outcome causal graph comparison and analysis (\cref{fig: workflow}). This section introduces the components and details of the functionality of our visual analysis system, describes the typical analysis processes, and reports on the implementation. 
The user interface of our visual analysis system is shown in~\cref{fig:interaction_process} and~\cref{fig:sgUI}.
Please refer to the supplemental video for the visual analysis process.

\subsection{Single Graph Analysis}
\label{sec:sinGraphExplo}
To realize our single causal graph exploration and analysis, we design the user interface of \emph{single outcome graph analysis} (\cref{fig:sgUI}), which consists of three key views: (1) Dataset and Variables Selection View (\cref{fig:sgUI}(1)), (2) Single Directed Graph View (\cref{fig:sgUI}(2)), and (3) Variables Matrix View (\cref{fig:sgUI}(3)).
A typical visual analysis of the causal graph of a single outcome is described as follows.

\begin{figure*}[tb]
    \centering
    \includegraphics[width=0.9\linewidth]{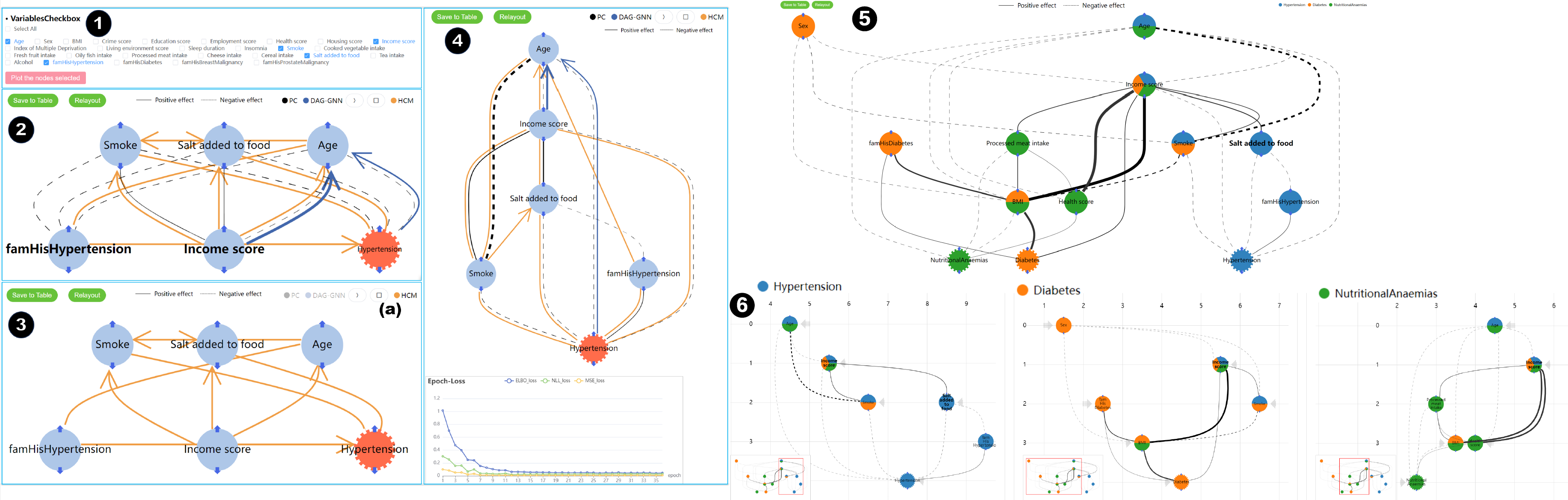}

    \caption{The case study of UKB data with a medical expert. Single causal graph analysis (1--4) and multi-outcome causal graphs comparison (5, 6).}
        \vspace{-0.3em}
    \label{fig:case_hypertension}
\end{figure*}

\subsubsection{Variable Selection}
\label{sec:variable_select}
The user should first choose the dataset (\cref{fig:sgUI}(a)) and then select the outcome of interest via a drop-down box (\cref{fig:sgUI}(b)).
To start with, the user can choose the top $n$ variables that have the highest correlation coefficients with the outcome via an input box (\cref{fig:sgUI}(c)), which helps reduce the complexity of the analysis to some extent (\textbf{R4}).
The user can then refer to the correlation plot (\cref{fig:sgUI}(d)) (blue for positive correlations and orange for negative) sorted by correlation coefficients of the selected outcome with all other variables in descending absolute value order to decide whether to add or delete some nodes for causal analysis.
Any variable of the dataset can be included or removed for the causal analysis by selections in the variable panel (\cref{fig:sgUI}(f)).

\subsubsection{Causal Exploration and Construction of Causal Graphs}
We consider the results calculated by the PC algorithm as the skeleton, and the results of DAG-GNN and HCM as references. 
Therefore, in our current design, all operations on nodes or edges are based on the results of PC, and users can refer to the results of the other two algorithms to manipulate the current single causal graph.

Variables selected in~\cref{sec:variable_select} are input to the three causal discovery algorithms, and an initial causal graph (\textbf{R1}) is obtained using the layered graph layout (\cref{sec:vis_causal_encode}) based on the result computed by PC. 
In the meantime, causal graphs calculated by DAG-GNN and HCM are updated progressively in real-time as the two algorithms run in the backend and are shown with our comparative visualization (example in \cref{fig:interaction_process}(1)) (\textbf{R2, R3}). 
Users can decide whether to pause, resume, or stop with the widgets (\cref{fig:interaction_process}(d)) to control the progress of DAG-GNN.

The superposition of edges of the three algorithms (\cref{fig:interaction_process}(e)) and the ``epoch-loss'' plot showing the model quality measures of DAG-GNN (\cref{fig:interaction_process}(f)) can improve the interpretability of causal discovery (\textbf{R2, R3}) and support the user to determine a reasonable causal relation (\textbf{R5}). 
For example, if an edge exists in all three results, it is likely a true causality.
The user can find the Variables Matrix View (\cref{fig:sgUI}(3)) that shows the multidimensional visualization of variables (including the outcome) obtained in the Dataset and Variables Selection View (\cref{fig:sgUI}(1)).
By analyzing the distributions, the user can gain more insights into the relationships between any pair of selected variables (\textbf{R4}).

Combining domain knowledge with insights gained from the visualizations, the user can modify the causal graph interactively (\textbf{R5}). 
The variables checkbox (\cref{fig:interaction_process}(a)) allows adding or removing nodes. 
By clicking on an edge, a menu pops up that allows deleting or reversing an edge (\cref{fig:interaction_process}(b)). 
By connecting the blue widget of one node to the blue widget of another node, 
the user can add an edge (\cref{fig:interaction_process}(c)).
A re-layout of the causal graph can be initiated after edits for better visualization.

After iterative interactions with all these functionalities, a more reasonable causal graph for the outcome of interest can be obtained.
The edited graph can be stored in the history for the visual comparison of multi-outcome causal graphs in the second stage.

\subsection{Multi-outcome Graphs Analysis}
The user repeats the process in~\cref{sec:sinGraphExplo} to obtain individual causal graphs for different outcomes and store them in the history panel (\cref{fig:interaction_process}(g)). 
After that, the user can select graphs of interest for comparison from the history panel (\textbf{R6}). 
By manipulating the different tabs, the user can visualize the multi-outcome graphs in various styles including the supergraph (\emph{Super Graph}),
our new comparable layout for subgraphs (\emph{Relative Subgraphs}), and subgraphs directly extracted from the supergraph (\emph{Extracted Subgraphs}) serving as an uncompressed reference to our method.
In all cases, each subgraph can be interactively manipulated with zooming and panning for in-depth comparisons. 
Aided by visualization, the user can gain insight into the similarities and differences between any two subgraphs (i.e., disease outcomes) regarding influencing factors and the extent of the effects of causality.

\subsection{Implementation}
Our visual analysis system is implemented using JavaScript and Python. 
The front-end functionality is realized with the Vue2 framework and data visualization libraries Echarts and D3.js. 
The graph layout computation is aided by the dagre library.
The causal analysis and data processing in the back-end are implemented in Python.
Multi-threading is used to support the progressive visualization of single causal graphs.
The causal effect estimations are calculated using the causal inference library DoWhy. 

\section{Case Study}
\label{sec:case_study}
We demonstrate the usability of our multi-outcome analysis system with a case study investigating the causality of various medical conditions.
We applied our approach to two health research datasets (permissions obtained),  CLHLS~\cite{DVN2020} (a longitudinal survey focusing on the health and longevity of elderly people in China, collecting a wide range of health, social and economic data) and UKB\footnote{\url{https://www.ukbiobank.ac.uk}} (a database of extensive biomedical data on approximately 500,000 UK residents, covering genetic, lifestyle, and health information).

We invited one domain expert (\textbf{E1}, a co-author who was involved in the design study) to participate in our case study hosted in our laboratory. 
\textbf{E1} is an assistant professor at a medical university with more than 10 years of clinical medicine experience. She holds a clinical medicine license and performs health data research. 
The study began with a comprehensive 20-minute presentation on the background of causal discovery methods, and our two-stage system design and visualization. 
Subsequently, the expert dedicated an hour to freely explore our system and expressed her thoughts through a think-aloud method. 
The observers listened and talked to the expert to answer any questions.
Finally, a 30-minute post-study interview was conducted to gather subjective comments, questions, and feedback from the expert on the method and the system.
High-resolution images of the case study can be found in the supplemental material.

\subsection{Causality in UKB}
We used the preprocessed original UKB dataset to obtain 8 frequently observed diseases as outcome variables and 27 influencing variables related to physiological condition, education level, lifestyle, diet, family history, etc., which may affect these outcomes. 
The final dataset we used in our research contains records of 98,530 people with at least one out of the 8 outcomes.
We report on two representative insights gained by the expert in this section.

\textbf{Single causal graph of hypertension.} The expert first chose hypertension (\cref{fig:sgUI}(b)) as the outcome and selected the top 5 (\cref{fig:sgUI}(c)) most relevant variables as potential influencing factors (\cref{fig:sgUI}(e)) to explore the causal relationship between these variables. 
Compared to ``Sex'' and ``Index of Multiple Deprivation'', she thought that ``Age'' and ``Salt added to food'' were more likely to affect the occurrence of hypertension, so she changed the nodes through the variables checkbox (\cref{fig:case_hypertension}(1)) and then recalculated the DAG (\cref{fig:case_hypertension}(2)). 
Users can switch on or off the legend of each algorithm (\cref{fig:case_hypertension}(a)) to show a specific result (results with HCM alone are shown in~\cref{fig:case_hypertension}(3)). 
Observing these three results, \textbf{E1} commented that ``Each algorithm can calculate causal relationships that are partially consistent with domain knowledge, but none is perfectly correct'', which supports our choice to apply multiple algorithms to complement each other. 
Combining domain knowledge and insights gained from the three algorithms, the expert modified the original DAG (\cref{fig:case_hypertension}(2)), and the final causal graph approved by the domain expert is shown in \cref{fig:case_hypertension}(4). 
Some causal relationships, such as the causal relationship of the negative effect value of ``Age $\rightarrow$ Hypertension'' and ``Smoke $\rightarrow$ Hypertension'', seem inconsistent with the first impression, however, \textbf{E1} considered it may be due to the reality that ``fewer heavy smokers can live old enough''.

\textbf{Causal comparison of multiple outcomes.} After analyzing the causal graph of hypertension, the expert further explored the causal graph of diabetes and nutritional anemia for further comparison (\cref{fig:case_hypertension}(5), (6)), as she anticipated that these different diseases might coexist in the same patient and tend to share some influencing factors, but the same factors have different mechanisms on different diseases.
Hypertension and diabetes share ``Income score'' and ``Smoke'', hypertension and nutritional anemia share ``Age'' and ``Income score'', diabetes and nutritional anemia share ``Income score'' and ``BMI''.  
\textbf{E1} found that a supergraph is easy for comparing two different outcomes. 
However, when three or more outcomes are compared, it becomes difficult with a supergraph (\cref{fig:case_hypertension}(5)), in which case our method can help with better comparisons (\cref{fig:case_hypertension}(6)).
Note the anchored shared nodes ``Income score'', ``Smoke'', and ``Age'' denoted with grids in~\cref{fig:case_hypertension}(6). 

For the shared node ``Income score'', positive and negative causal effect values of ``Income score $\rightarrow$ Hypertension'' (negative) and ``Income score $\rightarrow$ Diabetes'' (positive) are found. 
\textbf{E1} thought it is reasonable that people with a higher ``Income score'' have stronger health awareness, and therefore are less likely to develop hypertension. 
However, the other links that the wealthy have higher BMI and are more likely to develop diabetes are unusual; \textbf{E1} commented that ``It may be associated with the sampled people in this data'' and further investigation is needed. 

The effect of ``Income score'' on anemia is not direct. 
Smoking has a direct causal effect on the occurrence of hypertension, but the causal effect on the occurrence of diabetes is not quite clear, suggesting that smoking cessation behavior may be more beneficial to the control of hypertension, which is in line with the expectations of the expert.
Similarly, the causal relationships in the graph suggest that salt restriction may be better for hypertension control and weight loss may be better for diabetes control. 
Obese people (high BMI) are more likely to develop diabetes and less likely to develop nutritional anemia. 
Insights gained from the comparison of different outcomes are in line with the expectations of \textbf{E1} and are beneficial to guide the intervention in the prognosis of various diseases.

\subsection{Causality in CLHLS}
The CLHLS data was preprocessed to have 13,860 records with 10 outcome variables and 28 influencing variables related to physiological condition, education level, lifestyle, social and economic status, etc. 
We asked \textbf{E1} to explore this dataset for cross-validation using the same outcome variables, hypertension, and diabetes, as in the exploration of the UKB dataset described above.

\textbf{E1} got similar insights from this dataset.
For example, ``\( {f31\_sum} \)'' is an indicator of economic sources, and a higher value indicates a better income situation, and thus a lower possibility of having hypertension and diabetes; obese people (high BMI) are more likely to develop hypertension and diabetes, etc.

Interesting differences to the UKB data were also found: for example,
people in the CLHLS are older than those in the UKB, \textbf{E1} commented that ``it is likely that the older, the lower rate of hypertension and diabetes, since those with diseases will not live long enough''.

\section{Expert User Evaluation}
We invited three domain experts (\textbf{E1, E2, E3}) all with more than 8 years of clinical or public health experience to evaluate our system (\textbf{E2} and \textbf{E3} are not involved in the design study). 
Since the case study took a long time (more than 2 hours), \textbf{E2} and \textbf{E3} were only involved in the evaluation phase due to their tight schedule.
\textbf{E2} is an associate professor of public health, and \textbf{E3} is a doctoral student of health data science, both at the university that the first/corresponding author is affiliated with. \textbf{E2} studies cancer epidemiology, data science of epidemiology, and potential risk factors of common diseases, while \textbf{E3} has extensive research experience in chronic diseases.
The evaluation took place in our lab using a laptop computer extended with a 27-inch display. Two researchers (authors) first introduced the method and the tool and demonstrated the use of the tool with the CLHLS. The experts were then asked to explore the UKB data by themselves.
The think-aloud protocol was used during the evaluation: the experts made comments and asked questions while the researchers answered their questions.

All experts rated the two-stage analysis framework positively, as it allows them to first obtain the causality of one specific outcome and then compare those different yet correlated diseases, giving them insights to better understand conditions like multimorbidity or comorbidity and eventually help improve the prognosis. 
They agreed that the comparative visualization of different algorithms is beneficial for guiding the editing of causal graphs (\textbf{R2, R5}).
\textbf{E1} commented: ``If all three algorithms calculate an edge, I would be inclined to assume that there is a high probability that the edge exists, which helps to edit the causal relationship reasonably when there is doubt about some causal relationship.''
In addition, the progressive visualization of DAG-GNN improves the interpretability of the results (\textbf{R3}), according to \textbf{E2}: ``I will choose to pause the algorithm when the line chart of epoch-loss tends to be stable, which indicates that the algorithm has obtained relatively satisfactory results and can be referred to.''  
Experts commented that the visualization of multivariate variables (\cref{fig:sgUI}(3)) and correlation coefficients (\cref{fig:sgUI}(d, e)) are helpful in the selection of variables (\textbf{R4}). 

They also commented on the visualization and encoding design of our system. 
The layered layout from top to bottom 
makes it easy to intuitively determine the direction of causality (\textbf{R1}). 
All experts agreed that the visual encoding of nodes and edges is intuitive, and the operation process is relatively simple. 
\textbf{E3} commented that ``it is easy for me to edit the causal graph combining with domain knowledge'' (\textbf{R5}). 
Regarding multi-outcome causal graph comparison (\textbf{R6}), all experts agreed that it is convenient with the \emph{history view} (\cref{fig:interaction_process}(g)). 
They appreciated both forms of visual comparison: the supergraph (\cref{fig:interaction_process}(h)) and relative subgraphs (\cref{fig:interaction_process}(i)) for different situations.
They commented that the supergraph is preferred for up to three outcomes, while the relative subgraphs should be used for more outcomes.

\section{Discussion}

In our current analysis system, users are asked to select a limited number of potential variables to reduce the algorithm running time, partly supported by calculating the correlation coefficients. 
Note that the datasets we used contain both categorical and continuous variables, suggesting we should apply different methods to calculate the correlation coefficients~\cite{Schober2018}.
However, we argue that correlation coefficients calculated by different methods are incomparable and, therefore, confusing. 
Moreover, we only use the correlation coefficient to assist in the selection of variables, and the final selection of variables is more based on expert domain knowledge, therefore, we choose to use the Pearson correlation coefficient uniformly. Domain experts commented that providing more advanced representations of influencing factors would be useful, for example, by calculating multivariate logistic regression, or through data-driven predictive models instead of the correlation coefficients.
We will explore this possibility in future work.

Both CLHLS and UKB data used in our case are cross-sectional, which means that the direction of causality between variables cannot be fully determined. 
For example, people who smoke or are heavy on salt are more likely to develop hypertension, but in cross-sectional data, it is also possible that they are told to quit smoking and limit salt intake because of high blood pressure. 
If cohort data are available, the sequence of events in time can be used to further determine the causal direction. In addition, as mentioned in~\cref{sec:case_study}, our current analysis is based on partial variables extracted from two datasets. More interesting discoveries may be made if we can access more comprehensive data.

The computational scalability~\cite{Richer2022} of our method is mainly limited by the causal discovery methods.
We conducted a scalability study with varying number of variables for the CLHLS and UKB datasets. The study suggests that our method is feasible for the causal analysis of large datasets with less than ten variables per outcome with reasonable computational time.
Although we did not perform a formal visual scalability study, our experience shows that the visualization becomes cluttered if more than ten variables are involved per outcome.
Experts commented that focusing on the most important factors, typically less than ten, was sufficient to solve most of the specific domain problems.
For seven variables, PC took 6 seconds (for the CLHLS dataset) and 52 seconds (UKB dataset) to finish, DAG-GNN took 2 minutes (CLHLS) and 3 minutes (UKB) to be stable, and HCM took 3 minutes (CLHLS) and 4 minutes (UKB), all on a machine with Intel i7 2.60\,GHz CPU and 16\,GB RAM. To address the issue of long computation times for DAG-GNN and HCM, we designed our system to perform their computations progressively to avoid interference with user interactions. 
Details of the study are documented in the supplemental material.

Even though the method aims to resolve specific medical problems, several new techniques in our method address general issues in causal analysis and visual comparison. 
For example, causal discovery is enhanced by the synergy of various learning algorithms, including state-of-the-art techniques; the deep-learning-based causal discovery method is integrated into interactive visual analysis with progressive visualization that also improves interpretability; comparative visualizations enable effective comparisons of multiple causal graphs and multiple casual learning techniques. Our method is thus possible to be applied to other datasets that contain more than one outcome for multi-outcome comparison. 
For example, in business, purchasing decisions, brand loyalty, shopping experience, etc.\ can be regarded as different outcomes, and there may be the same or different influencing factors among these outcomes. Analyzing the causal differences between these outcomes can help enterprises optimize marketing strategies and improve product design according to various goals.

We attempted to compare our method with existing visual analysis methods of causal graphs~\cite{Xie2021,Wang2016,Wang2017,Hoque2022}.
However, none of them provides readily usable tools or source codes.
Therefore, we could only refer to the papers and associated video demos.
After examining these sources, we believe that these methods can partly address the single outcome causal graph analysis problem but do not support the creation and the subsequent comparative analysis of multiple causal graphs with different outcome variables.
Moreover, each of these works uses only one classical causal discovery method, e.g., GES, F-GES, PC, to generate causal graphs.

\section{Conclusion and Future Work}
In conclusion, we have introduced the visual analysis of multi-outcome causal graphs that were previously not studied in visual analytics.
Inspired by a domain problem in healthcare, the method is designed in collaboration with medical experts.
Our method consists of two stages: starting from a single outcome, analysts explore, analyze, and edit the associated causal graph discovered by a combination of causal discovery algorithms aided by our progressive visualization technique; causal graphs of single outcomes are then visually compared using a new graph layout and visual mappings as line-ups or as a supergraph.
The usefulness of our method was evaluated through a case study and an expert user study with medical experts using widely accepted medical datasets.

In the future, we would like to improve the guidance of initial variable selection using more advanced data analysis techniques than correlation analysis. 
We would also like to support the analysis of causal graphs with external evidence in addition to the data, such as well-accepted knowledge graphs and conclusions learned from state-of-the-art publications on the domain problem.
Another direction of improvement is to study the temporal aspect of cohort data as it provides vital information for understanding causality, especially in healthcare problems.

\section{Supplemental Material}
\label{sec:supplement_inst}
We include content left out for conciseness from the paper in this supplemental material. Specifically, we include more details on the three selected causal discovery algorithms, numerical measurements of these algorithms on two benchmark datasets, our comparable graph layout algorithm, the results of the computational scalability study, and high-resolution figures of the paper in larger sizes.

\subsection{Causal Discovery}
\label{sec:causalGraphMethod}
In this section, we introduce the causal discovery algorithms used in our method and report on numerical measurements of these methods.
\subsubsection{Causal Discovery Methods}
Three causal discovery methods are adapted into our system: the classic Peter-Clark (PC) algorithm~\cite{Spirtes1993}, which is known to provide a good causality skeleton, the DAG-GNN method~\cite{Yu2019} as a representative modern continuous optimization method that can handle mixed-type data, and the hybrid causal learning method (HCM)~\cite{Li2022}, specifically designed for mixed-type data.

(1) PC

The PC~\cite{Spirtes1993} algorithm is a well-known and frequently-used causal discovery method that serves as the backbone approach for quickly analyzing the selected variables. The method starts with a fully connected undirected graph and identifies a skeleton by evaluating the conditional independence of each variable pair $X_i \indep X_j |\mathbf{S}$, where $\mathbf{S}$ is the conditioning set. Next, the edges are oriented by identifying colliders/v-structures, e.g., $X_i\rightarrow X_k \leftarrow X_j$ in the skeleton. The algorithm then orients additional edges that would otherwise form more v-structures.

(2) DAG-GNN

The original combinatorial optimization problem can be formulated as finding a score function $s(\mathbf{A})$ for a DAG described by an adjacent matrix $\mathbf{A}$ of $d$ variables as the left-hand side of ~\cref{eqn:contFormations}.
As a continuous optimization method, DAG-GNN~\cite{Yu2019} solves the transformed continuous optimization problem formulated by the NO TEARS method~\cite{Zheng2018} as the right-hand side of ~\cref{eqn:contFormations}.
\begin{equation}
\begin{aligned}[c]
&\textstyle \min_{\mathbf{A}\in \mathbb{R}^{d\times d}}s(\mathbf{A})\\
&\text{subject to\space} G(\mathbf{A}) \in \text{DAGs}
\end{aligned}
\qquad\Longrightarrow\qquad
\begin{aligned}[c]
&\textstyle \min_{\mathbf{A}\in \mathbb{R}^{d\times d}}s(\mathbf{A})\\
&\text{subject to\space} h(\mathbf{A}) = 0 \;.
\end{aligned}
\label{eqn:contFormations}
\end{equation}
Here, the penalty function $h(\mathbf{A})$ enforces acyclicity:
\begin{equation}
    h(\mathbf{A}) = \text{trace}(e^{\mathbf{A}\odot \mathbf{A}}) - d = 0\;,
\end{equation}
where $\odot$ is the Hadamard product. 

DAG-GNN solves the optimization problem using a generative graph neural network based on variational autoencoders with neural network functions and evidence lower bound.
It extends the NO TEARS method for nonlinear structural equation models and supports both discrete and continuous variables.

However, as a deep learning method, DAG-GNN takes a long time to train and it is not feasible for the user to wait until the training to finish in a visual analytics environment.
Therefore, we enable the progressive visualization of intermediate results of DAG-GNN and support interactive pausing/resuming of the algorithm in our method.

(3) HCM

The HCM is a recent advancement for mixed-type data.
It addresses the mixed-type data problem by formulating a mixed structure equation model~\cite{Li2022}.
For continuous variables $X_j \in \mathbb{R}$, or categorical variables $X_j \in \{1,\cdots,c_j\}$, the mixed structure equation model reads:
\begin{equation}
\begin{cases}
 X_j = f_j (pa_j) + u_j\;,&\text{for continuous variables,}\\
 X_j = \mathrm{arg\,max}_{k\in\{1,\cdots,c_j \}}f_{j,k}(pa_j) + u_{j,k}\;,&\text{for discrete variables}\;.
 \end{cases}
\end{equation}
Here, $pa_j$ indicates the parents of $X_j$.
The model is learned with skeleton learning using the PC-stable algorithm~\cite{Colombo2014} with a new mixed-type conditional independence test; then, the method finds causal graph directions via greedy search with a mixed information criterion; and the graph is further pruned using the conditional independence test.
Note that the skeleton learning step can be replaced by other methods, for example, the aforementioned traditional PC algorithm.

\subsubsection{Numerical Measurements on Benchmark Datasets}
The numerical measurements of the causal discovery methods on two benchmark datasets: healthcare\footnote{\url{https://www.bnlearn.com/bnrepository/clgaussian-small.html\#healthcare}} and prepd-bathia\footnote{\url{https://www.bnlearn.com/research/scirep17/}}, are summarized in~\cref{tab:causalMeasure}.
Here, we record typical measurements in causal discovery, including accuracy, false positive rates (FPR), and Hamming distance (Hamming dist.). 
\begin{table}[htb]
\scriptsize
    \caption{Numerical measurements of causal discovery methods. H:~``healthcare'' dataset, P:~``prepd-bathia'' dataset. }
    \label{tab:causalMeasure}
    \centering
    \begin{tabu}{cccc}
       \toprule
        Method& Accuracy (H, P) & FPR (H, P) & Hamming dist. (H, P) \\
        \midrule
            PC & 0.67, 0.67 & 0.42, 0.22 & 8, 6\\
            DAG-GNN & 0.56, 0.33 & 0.33, 0.46 & 8, 19 \\
            HCM & 0.67, 0.75 & 0.17, 0.08 & 5, 5\\
            Combined & 1.00, 0.92 & 0.67, 0.54 & 8, 14\\
        \bottomrule
    \end{tabu}
\end{table}
\subsection{Comparable Layout for Multiple Causal Graphs}
We provide more information about our comparable layout for multiple causal graphs, including a detailed description of the layout algorithm, and the comparison of the graph layout of direct extraction and the comparable graph layout.

\subsubsection{Comparable Layout Algorithm}
The objective function for a given subgraph $G_t$ of nodes $V$ with input coordinates $\mathbf{x}$ in the supergraph and output positions $\mathbf{\xi}$ reads:
\begin{align}
    \min_{V\in G_t}& \sum_{i,j\in Q}w_i(||\mathbf{x}_i - \mathbf{x}_j|| - ||\mathbf{\xi}_i - \mathbf{\xi}_j||)^2 +\nonumber\\
    & \sum_{k,l\in C_t}w_l(||\mathbf{x}_l - \mathbf{x}_k|| - ||\mathbf{\xi}_l - \mathbf{\xi}_k||)^2  \label{eqn:layoutGoal}\\
    \text{subject to: } & V \; \text{do not violate ranks and orders} \nonumber\\
    & V \in V_Q\; \text{are anchored} \;.    \nonumber
\end{align}
Here, $Q$ is the set of indices of all common nodes $V_Q$ shared by any subgraphs, and $C_t$ is the index set of unique nodes of $G_t$.

We detail the algorithm of the comparable graph layout in~\cref{alg:layout}.
\begin{algorithm}[htb]
\caption{Horizontal compression of subgraphs}\label{alg:layout}
\begin{algorithmic}
\Require $G_s$ has the  optimal layout
\Ensure $G_t$ are compact and relative node pos of $G_s$ are preserved
\State $V_Q \gets$ nodes shared by more than one graph
\State $L(V_Q) \gets $Sort the unique horizontal coords of $V_Q$ \Comment original coordinates 
\State $L_\xi(V_Q) \gets L(V_Q)[1]$+[0,1,$\ldots$, length($L(V_Q)$)-1] \Comment output coordinates
\For{$t=1,2,\ldots, n$} \Comment{for all outcomes}

\State $V_t \gets $ unique nodes of $G_t$
\State $V_Q^t \gets $ shared nodes of $V_Q$ in $G_t$
\State $L(V_t) \gets $Sort the unique horizontal coords of $V_t$
\State $N(V_t) \gets $ record \# of unique nodes between elements of $L(V_Q)$
\State $D(V_t) \gets $ unit distances based on node relationships and $N(V_t)$ \Comment node distance list of unique nodes
\State $L_\xi(V_t) \gets$ adjust positions of $L(V_t)$ by $D(V_t)$
\State $L_\xi(V) \gets $ merge $L_\xi(V_t)$ and $L_\xi(V_Q^t)$ sorted \Comment output coordinates
\State $V.x \gets$ $L_\xi(V)$ 
\EndFor
\end{algorithmic}
\end{algorithm}
First, we identify any nodes that are shared by two or more subgraphs in the supergraph to create the set $Q$ and compress them horizontally.
Nodes $ V_\mathbf{Q}$ are compressed by
assigning their horizontal coordinates ($L_\xi(V_Q)$) to a series of integers starting with a step size of 1---the common nodes are therefore anchored. 
Next, for a subgraph $G_t$, we find its unique nodes $V_t$ and shared nodes $V_Q^t$ in $Q$, and then compress their horizontal distances.
The horizontal coordinates of $V_t$ are sorted first, and then, 
the positions are adjusted to give the compressed unique node list $L_\xi(V_t)$.
Specifically, the adjustment is based on a unit distance list $D(V_t)$.
Different unit distances are used depending on the relationship between the unique node and common nodes: if the horizontal coordinate of the unique node is smaller than the smallest common node or is greater than the largest common node, the unit distance is 1; otherwise, the unit distance is $1/(N(V_t)+1)$---unique nodes are evenly distributed between two common nodes.
The final horizontally compressed node list $L_\xi(V)$ is created by merging $L_\xi(V_Q^t)$ and $L_\xi(V_t)$.

\subsubsection{Visualization of Multi-outcome Causal Graphs}
\Cref{fig:causal_comparison} shows the comparison of subgraphs (a) laid out using direct extraction from the supergraph method and (b) with our new comparable graph layout.
The horizontal stress is labeled as ``stress(X)'' on top of each graph.

\subsection{Computational Scalability}
The computational scalability results of our method from the scalability study are summarized in~\cref{tab:scalability}.
The CLHLS dataset~\cite{DVN2020} and the UKB dataset\footnote{\url{https://www.ukbiobank.ac.uk}} contains 13,860 records with 28 influencing variables, and 98,530 records with 27 influencing variables, respectively.
Tests were performed on a PC with an Intel i7 2.60\, GHz CPU and 16\, GB RAM running Windows~11. 
\begin{table}[htb]
    \caption{%
      	The approximate running time of the three causal discovery algorithms under different numbers of selected influencing factors of two datasets.
      }
    \label{tab:scalability}
    \scriptsize%
    \small
    \centering%
    \begin{tabular}{ccccccc}
    \toprule
    \multirow{2}{*}{\textbf{\begin{tabular}[c]{@{}c@{}}\# of selected\\ influencing factors\end{tabular}}} & \multicolumn{3}{c}{\textbf{\begin{tabular}[c]{@{}c@{}}CLHLS\\ 13860*28\end{tabular}}} & \multicolumn{3}{c}{\textbf{\begin{tabular}[c]{@{}c@{}}UKB\\ 98530*27\end{tabular}}} \\
    \cmidrule(lr){2-4}\cmidrule(lr){5-7}& \textbf{7}                 & \textbf{14}                 & \textbf{28}                & \textbf{7}                & \textbf{14}                & \textbf{27}                \\
    \midrule
    \textbf{PC}                                                                                            & 6\,s                         & 10\,min                       & 4\,h                         & 52\,s                       & 1\,h                         & NA                         \\
    \textbf{DAG-GNN}                                                                                       & 2\,min                       & 3\,min                        & 10\,min                      & 3\,min                      & 6\,min                       & NA                         \\
    \textbf{HCM}                                                                                           & 3\,min                       & 7\,min                       & 40\,min                      & 4\,min                      & 10\,min                       & NA                         \\
    \bottomrule
    \end{tabular}
    \\
    \textsuperscript{*} The running times for computing 27 influencing variables with all three algorithms are marked as ``NA'' as they are excessively long. The running time of DAG-GNN refers to the approximate time when the algorithm becomes stable.
\end{table}

For the CLHLS dataset, when the number of selected influencing factor variables is 7, 14, and 28, the DAG-GNN algorithm tends to stabilize after running for 72, 136, and 298 epochs, respectively. 
For the UKB dataset, when the number of selected influencing factor variables is 7 and 14, the DAG-GNN algorithm tends to stabilize after running for 30 and 40 epochs, respectively.

\subsection{High-resolution Figures}
Figures 1, 4, and 7 of the paper are shown here as Figures~\ref{fig:teaser},~\ref{fig:sgUI}, and~\ref{fig:casestudy} at larger sizes for better visibility.

\acknowledgments{%
This work was supported in part by the NSFC grant (No.~62372012), the National Key R\&D Program of China (No.~2022YFC2705104), and the Deutsche Forschungsgemeinschaft (DFG, German Research Foundation) -- Project-ID 449742818. 
}

\bibliographystyle{abbrv-doi-hyperref}

\bibliography{multioutcomeCausality}

\newpage
\begin{landscape}
\begin{figure}[htb]
  \centering
  \subfloat[]{%
\includegraphics[width=\linewidth]{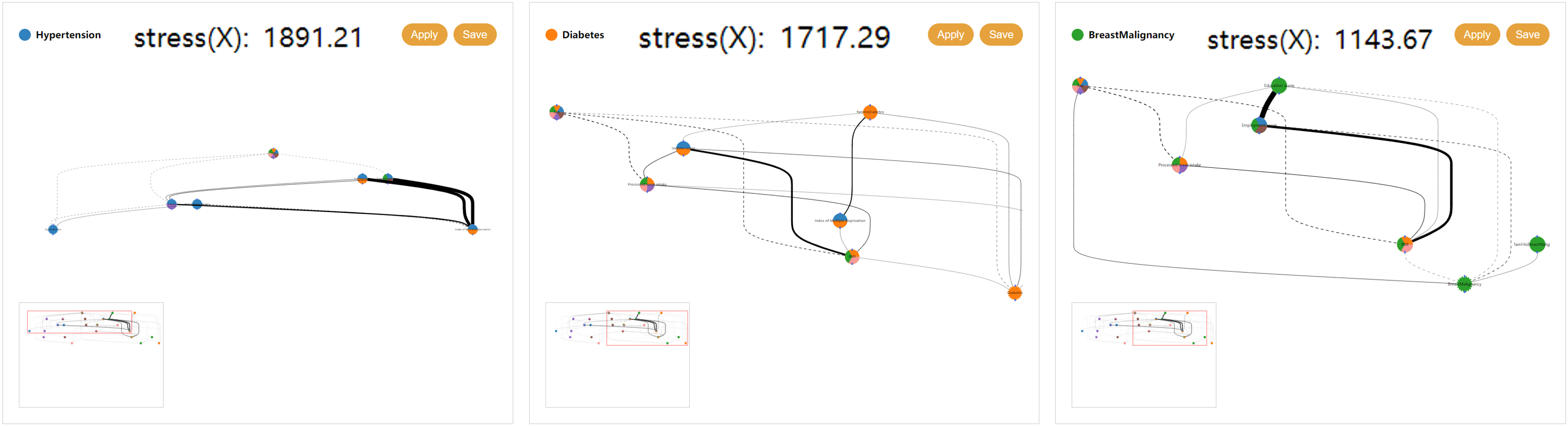}%
    \label{fig:three_outcome_extracted}%
  }\\%
  \subfloat[]
  {%
    \includegraphics[width=\linewidth]{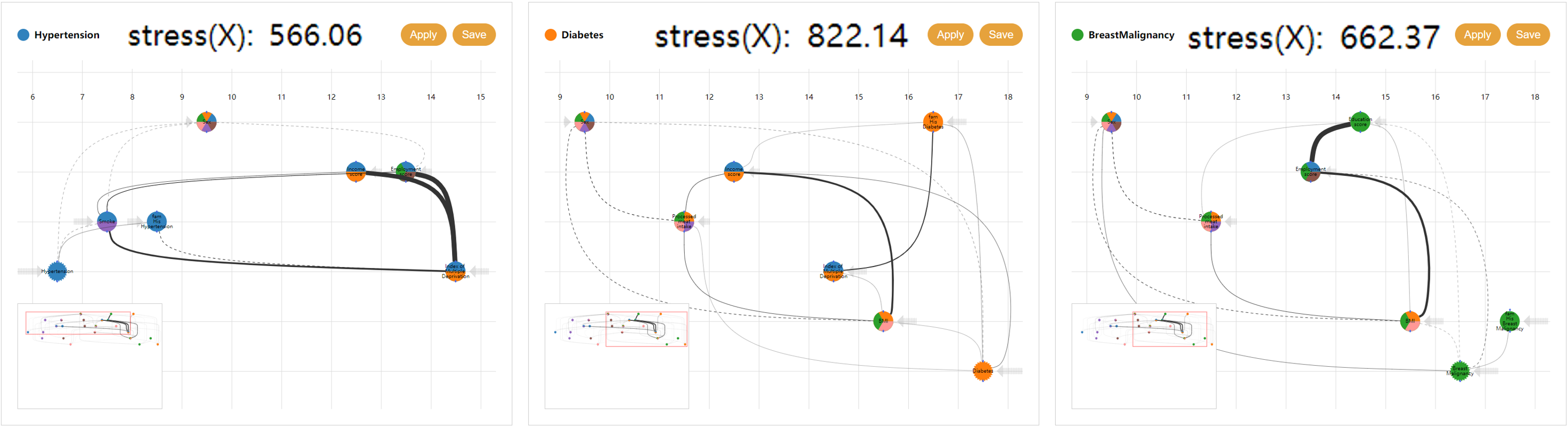}%
    \label{fig:three_outcome_relative}%
  }
  \vspace{-1em}
  \caption{Visualization of multi-outcome causal graphs
  with (a) subgraphs extracted directly from the supergraph and (b) our new comparable graph layout. 
  }
  \label{fig:causal_comparison}
\end{figure}    
\end{landscape}

\begin{landscape}
\begin{figure}[tb]
   \centering
    \includegraphics[width=\linewidth]{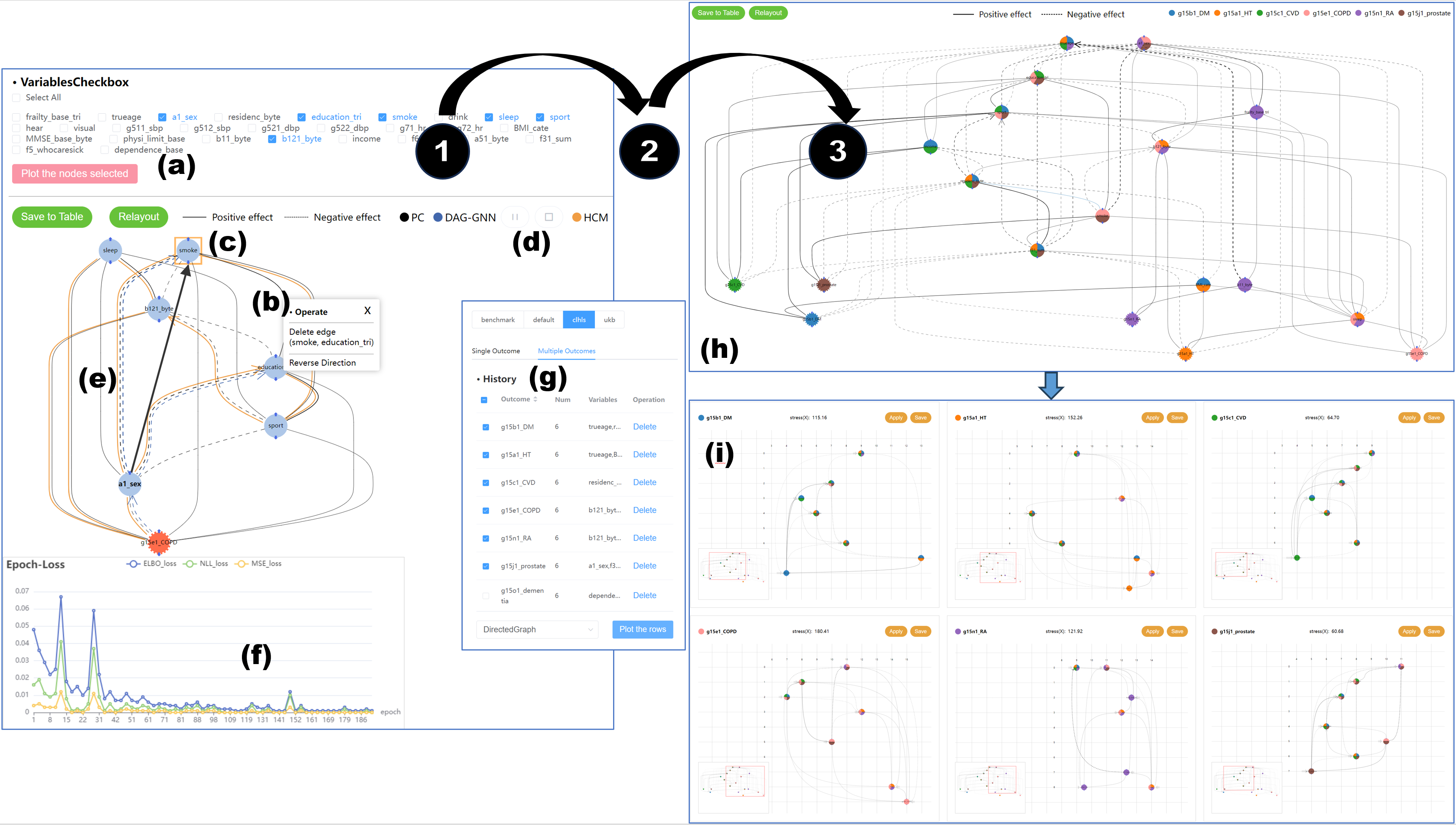}
\caption{The overall process of the visual analysis of multi-outcome causal graphs of a health research data~\cite{DVN2020}. Analysts use (1) the Single Graph View to analyze and edit the causal graph (a--c) of a single outcome aided by a set of new visualization techniques (d--f) to leverage the synergy of various causal discovery methods. The fine-tuned causal graphs of single outcomes are stored in (2) the History View (g). Next, causal graphs of interest are selected for visual comparison in (3) the Multi-outcome Graphs Comparison View (h, i) supported by our new graph layout and visual mappings (i).}
    \label{fig:teaser}
\end{figure}
\end{landscape}

\begin{landscape}
\begin{figure}[tb]
        \centering
    \includegraphics[width=\linewidth]{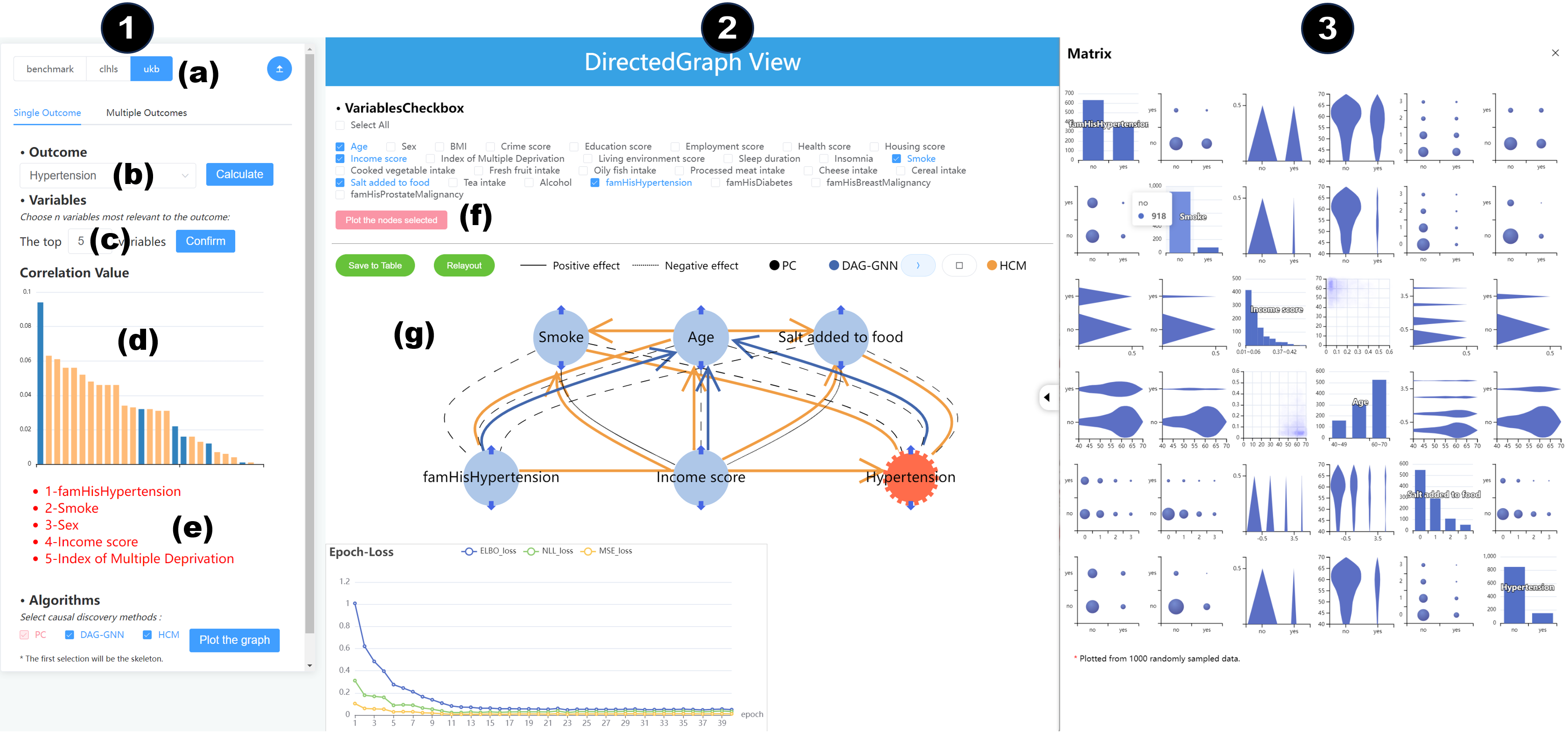}

    \caption{The user interface of our visual analysis system for single outcome graphs consists of three views: (1) Dataset and Variables Selection View allowing for dataset (a) and variables selection (b, c) and correlation analysis (d, e), (2) Single Directed Graph View allowing for graph editing (f, g), and (3) Variables Matrix View providing insight for graph editing.
    }
    \label{fig:sgUI}
\end{figure}
\end{landscape}

\begin{landscape}
\begin{figure}[tb]
        \centering
    \includegraphics[width=\linewidth]{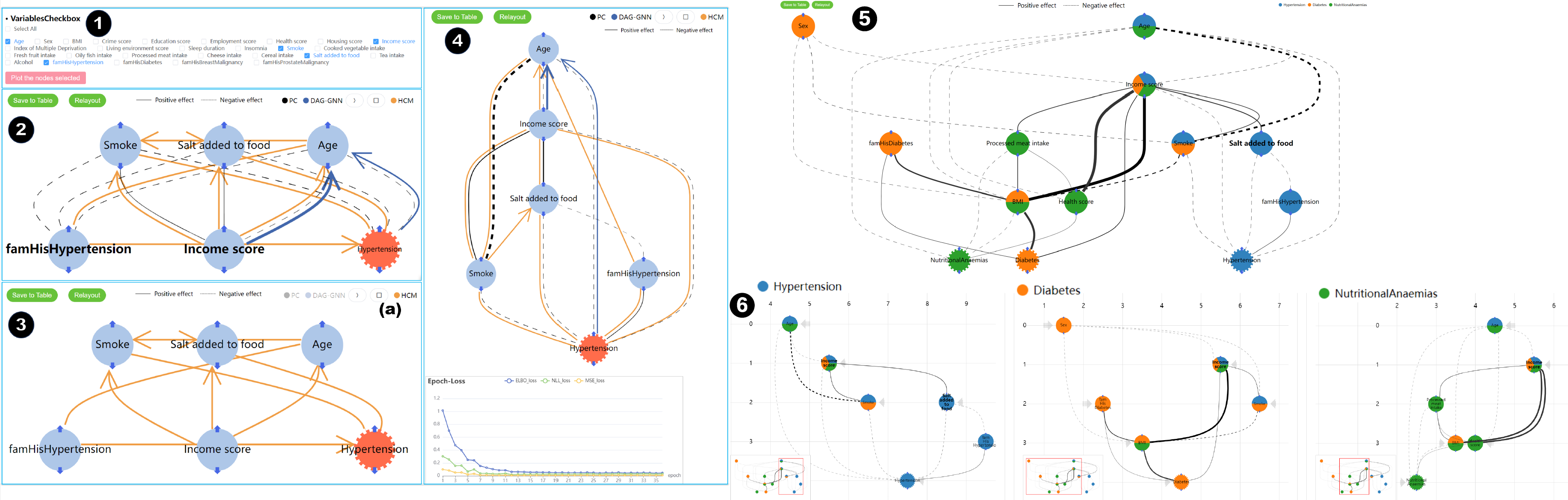}

    \caption{The case study of UKB data with a medical expert. Single causal graph analysis (1--4) and multi-outcome causal graphs comparison (5, 6).}
    \label{fig:casestudy}
\end{figure}
\end{landscape}

\end{document}